\documentclass{article}

\PassOptionsToPackage{numbers, compress}{natbib}

\usepackage[main, final]{neurips_2026}

\usepackage{microtype}
\usepackage{graphicx}
\usepackage{booktabs} %
\usepackage{calc}
\usepackage{hyperref}
\usepackage{notation_samy}
\newcommand{\MNISTAdd}{\textsc{MNISTAdd}}

\newlength\myheight
\newlength\mydepth
\settototalheight\myheight{Xygp}
\usepackage{tikz}
\usetikzlibrary{shapes.geometric} %

\DeclareRobustCommand{\marker}[2]{%
  \tikz[baseline=-0.5ex]{%
    \node[
      draw=#1!60!black, %
      fill=#1,          %
      shape=#2,         %
      inner sep=1.5pt,  %
      line width=1pt  %
    ] {};%
  }%
}

\newcommand{\approxhat}[1]{\tilde{#1}}

\newcommand{\CTSketch}{\textsc{CTSketch}\xspace}
\newcommand{\sumfour}{\text{sum}_4}
\newcommand{\apricot}{\textsc{APRICOT}\xspace}

\newcommand{\SemStrengthening}{\textrm{Semantic Strengthening}\xspace}

\newcommand{\TTround}{\textsc{TT-Round}\xspace}
\newcommand{\HaTT}{\textsc{HaTT}\xspace}
\newcommand{\TCI}{\textsc{TCI}\xspace}
\newcommand{\RSI}{\textsc{RSI}\xspace}
\newcommand{\TTSVD}{\text{TT-SVD}\xspace}
\newcommand{\wmc}{\operatorname{WMC}}
\newcommand{\wmchat}{\operatorname{W}\widehat{\operatorname{M}}{\operatorname{C}}}
\DeclareMathOperator{\KC}{KC}
\DeclareMathOperator{\MUL}{MUL}
\DeclareMathOperator{\RMUL}{ROUNDMUL}

\usepackage[capitalize,noabbrev]{cleveref}
\usepackage{amsmath}
\usepackage{amssymb}
\usepackage{mathtools}
\usepackage{amsthm}
\usepackage{booktabs}       %
\usepackage{amsfonts}       %
\usepackage{nicefrac}       %
\usepackage{microtype}      %
\usepackage{dsfont}
\usepackage{algorithm}
\usepackage{algpseudocode}
\usepackage{caption,subcaption}
\usepackage{stmaryrd} 
\usepackage{bbm}
\usepackage{xspace}
\usepackage[dvipsnames]{xcolor}
\usepackage{lipsum}
\usepackage{tabularx}
\usepackage[export]{adjustbox} %
\usepackage{makecell}
\usepackage{tcolorbox}
\usepackage{threeparttable}

\definecolor{exactcolor}{HTML}{ff8001} %
\definecolor{indepcolor}{HTML}{fb3299} %
\definecolor{factorcolor}{HTML}{028081} %
\definecolor{prodcolor}{HTML}{800080} %
\definecolor{kccolor}{HTML}{e9af2f} %

\hypersetup{
colorlinks=true,linkcolor=brown,citecolor=brown,urlcolor=orange
}

\newcommand{\hlexact}[1]{\textcolor{exactcolor}{#1}}
\newcommand{\hlkc}[1]{\textcolor{kccolor}{#1}}
\newcommand{\hlindep}[1]{\textcolor{indepcolor}{#1}}
\newcommand{\hlfactor}[1]{\textcolor{factorcolor}{#1}}
\newcommand{\hlprod}[1]{\textcolor{prodcolor}{#1}}
\newcommand{\hlexactb}[1]{\textcolor{exactcolor}{\textbf{#1}}}
\newcommand{\hlindepb}[1]{\textcolor{indepcolor}{\textbf{#1}}}
\newcommand{\hlprodb}[1]{\textcolor{prodcolor}{\textbf{#1}}}
\newcommand{\hlkcb}[1]{\textcolor{kccolor}{\textbf{#1}}}
\newcommand{\hlfactorb}[1]{\textcolor{factorcolor}{\textbf{#1}}}
\newcommand{\indep}{\hlindep{\text{I}}}
\newcommand{\fac}{\hlfactor{\text{F}}}
\newcommand{\prodap}{\hlprod{\text{P}}}

\renewcommand{\paragraph}[1]{\textbf{#1.}}
\makeatletter
\def\mf@color@e{exactcolor}
\def\mf@color@i{indepcolor}
\def\mf@color@f{factorcolor}
\def\mf@color@p{prodcolor}
\def\mf@up@e{E}
\def\mf@up@i{I}
\def\mf@up@f{F}
\def\mf@up@p{P}
\newcommand{\methodflags}[1]{%
  \textsuperscript{%
    \@tfor\mf@letter:=#1\do{%
      \textcolor{\csname mf@color@\mf@letter\endcsname}%
        {\csname mf@up@\mf@letter\endcsname}%
    }%
  }%
}
\makeatother

\theoremstyle{plain}
\newtheorem{theorem}{Theorem}[section]

\newtheorem{corollary}[theorem]{Corollary}
\theoremstyle{definition}
\newtheorem{definition}[theorem]{Definition}

\newtheorem{example}[theorem]{Example}
\theoremstyle{remark}

\title{Where to Approximate in Neurosymbolic Inference?}

\author{%
    Samy Badreddine$^{1,2,3}$ \quad Emile van Krieken$^{4}$ \quad 
     Luciano Serafini$^{2,3}$ \quad Antonio Vergari$^{5}$\\ %
    $^1$Sony AI \quad $^2$Fondazione Bruno Kessler \quad $^3$University of Trento \\
    $^4$ Vrije Universiteit Amsterdam \quad $^5$University of Edinburgh\\
  \texttt{samy.badreddine@sony.com} \quad
  \texttt{e.van.krieken@vu.nl} \\
  \texttt{luciano.serafini@fbk.eu} \quad
  \texttt{avergari@ed.ac.uk}
}

\begin{document}

\maketitle

\begin{abstract}
Probabilistic neurosymbolic methods rely on weighted model counting (WMC) to combine neural predictors with symbolic constraints.
Exact WMC is \#P-hard and typically scales poorly, so many methods resort to approximations.
By unifying existing approaches under a three-step bottom-up \emph{approximate compilation} pipeline, we study \emph{where} approximation budget is best allocated.
Inspired by tensor networks, we instantiate two steps with tensor train decompositions and derive new pipelines that exactly multiply approximated factors and recompress their products via SVD- or interpolation-based schemes.
Across six tasks in Sudoku, arithmetic reasoning, and scene understanding, and across increasing problem sizes, our pipelines produce WMC estimates orders-of-magnitude more accurate than prior methods.
This translates into stronger downstream learning on tasks where reasoning relies on tight interactions between symbolic factors, where cruder approximations ignore these dependencies.
Our pipelines open novel design space for effective approximate compilation of challenging probabilistic reasoning tasks.
\end{abstract}

\section{Introduction}
\label{sec:intro}
Probabilistic neurosymbolic (NeSy) systems combine neural perception with symbolic constraints, to enable learning from weak
supervision~\cite{manhaeve2018deepproblog,xu2018semantic} or to inject
symbolic knowledge into structured prediction
pipelines~\cite{ahmed2022semantic}.
Inference involves solving the \emph{weighted model count} (WMC) problem \citep{chavira2008wmc}:
computing the probability that a neural distribution over discrete concepts
satisfies a constraint.
Computing the WMC exactly is \#P-hard in general~\cite{valiant1979enumeration}.
One approach for solving practical instances tractably is \emph{circuit compilation} which compiles the constraint into a logical circuit, a process also called \textit{knowledge compilation} \citep{darwiche2002knowledge}, while the neural distribution is turned into a probabilistic circuit \citep{vergari2019tractable,choi2020pc,ahmed2022semantic}.
Then, under certain structural properties, these circuits can be efficiently multiplied and summed over to yield the exact WMC \citep{shen2016tractable,vergari2021compositional}. 
The same structural properties and multiplication operation are also behind popular bottom-up strategies for knowledge compilation \citep{pipatsrisawat2008new,darwiche2011sdd}.

However, compilation is itself hard, and exactly multiplying several compiled circuits blows up exponentially in the number of factors \citep{vergari2021compositional}.
This implies that compilation strategies can hit a wall on many real-world NeSy problems.
Even compiling a $9\times 9$ Sudoku board, often used in benchmarks~\cite{augustine_2022_visudopc,wang_2019_satnet}, as a constraint is out of reach for all exact compilers.
Without surprise, many NeSy methods focused on approximate computation of the WMC \cite{vankrieken2022anesi,smet2023differentiable,choi_ctsketch_2025,dierckx2025learning}.
However, these approaches have carried over with different perspectives and from different communities.
As such, the problem of understanding \emph{where} and \textit{when} to approximate in a systematic way is still open.

\begin{figure}[!t]
    \centering
    \setlength{\tabcolsep}{2.5pt} 
    
    \begin{minipage}[c]{0.51\textwidth}
        \centering
        \footnotesize
        \begin{tabularx}{\textwidth}{lXXX}
            \toprule
            Baseline & SI & SII & SIII \\
            \midrule
            Exact methods~\cite{darwiche2011sdd,xu2018semantic}
             & \hlkcb{KC}
             & \hlexactb{\textsc{mul}}
             & -- \\
            Fuzzy methods~\cite{badreddine2020ltn,vankrieken2022fuzzy} & -- & -- & \hlindepb{\textsc{indep}} \\
             Sem Strength~\cite{ahmed2023semstrenghtening} & \hlkcb{KC} & \hlexactb{\textsc{mul}} & \hlindepb{\textsc{indep}} \\
            \CTSketch~\cite{choi_ctsketch_2025} & \hlfactorb{\TTSVD} & -- & \hlindepb{\textsc{indep}} \\
            \apricot A & \makecell[l]{\hlexactb{\textsc{premul}},\\ \hlfactorb{\TTSVD}} & \hlexactb{\textsc{mul}} & \hlindepb{\textsc{indep}} \\
            \apricot B & \makecell[l]{\hlexactb{\textsc{premul}},\\ \hlfactorb{\TTSVD}} & \makecell[l]{\hlexactb{\textsc{mul}},\\ \hlprodb{\textsc{roundmul}}} & \hlindepb{\textsc{indep}} \\
            \apricot C & \makecell[l]{\hlexactb{\textsc{premul}},\\ \hlfactorb{\TTSVD}} & \makecell[l]{\hlexactb{\textsc{mul}},\\ \hlprodb{\textsc{roundmul}}} & -- \\
            \bottomrule
        \end{tabularx}
    \end{minipage}
    \hfill
    \begin{minipage}[c]{0.48\textwidth}
        \centering
        \includegraphics[width=\linewidth, valign=c]{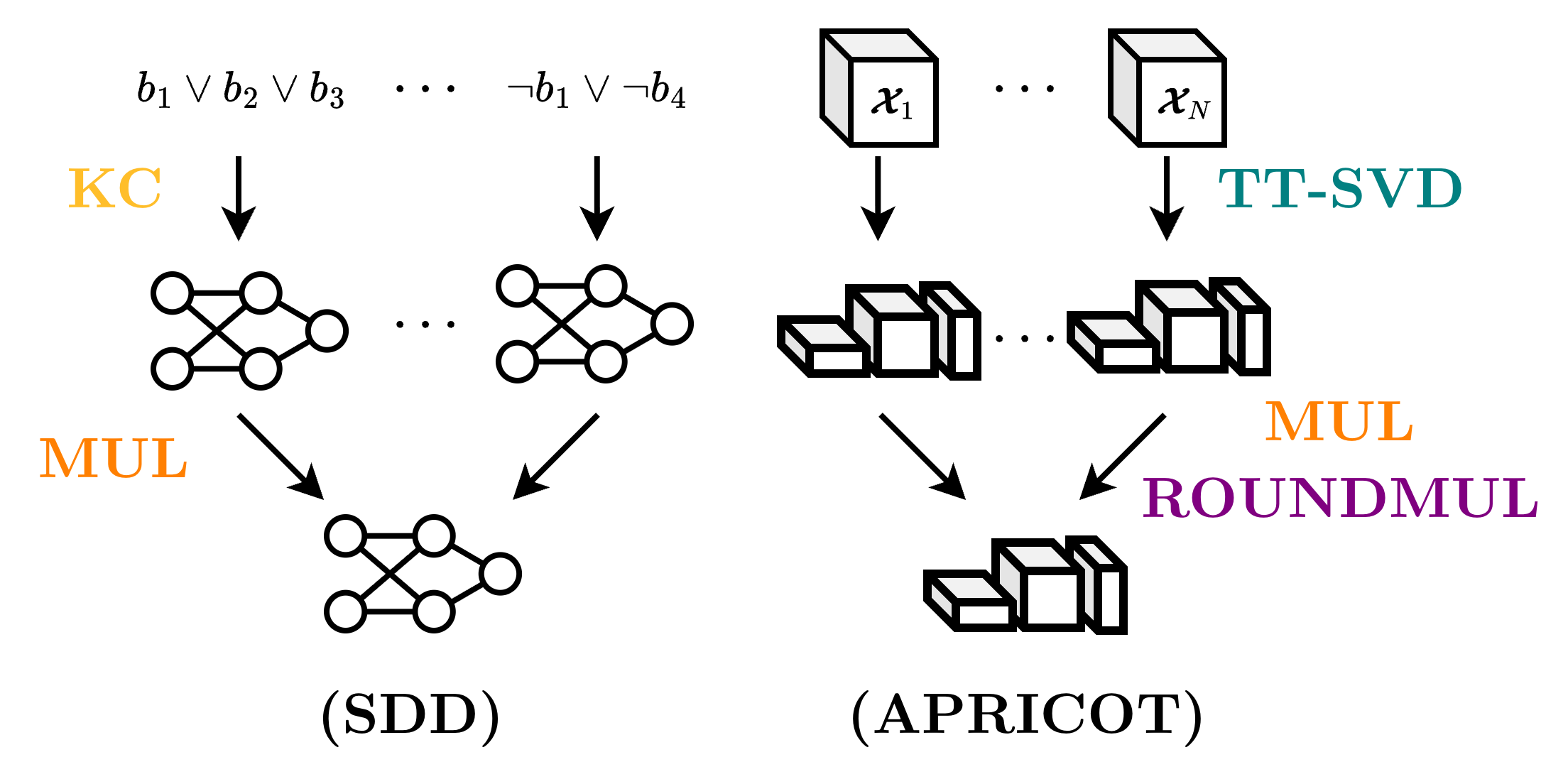}
    \end{minipage}
    \par\vspace{0.6em}
    {\centering\footnotesize
        \text{Exact operations:} \ 
        \textcolor{kccolor}{\rule{0.8em}{0.8em}} \ 
        \textcolor{exactcolor}{\rule{0.8em}{0.8em}}
        \hspace{2em}
        \text{Approximate operations:}\ 
        \textcolor{factorcolor}{\rule{0.8em}{0.8em}} \ 
        \textcolor{prodcolor}{\rule{0.8em}{0.8em}} \ 
        \textcolor{indepcolor}{\rule{0.8em}{0.8em}}
    \par}
    \caption{
    \textbf{Where can we approximate} when compiling structured-decomposable circuits?
    We unify bottom-up approaches in a three-step pipeline: \textbf{(Step I)} compiles each factor $\phi_j$ into an  \hlkc{exact} or \hlfactorb{relaxed} circuit,
    \textbf{(Step II)} multiplies the resulting $N$ circuits down to $K$ \hlexactb{exactly} or using \hlprodb{recompression strategies}, and \textbf{(Step III)} combines any $K>1$ remaining circuits with an \hlindepb{independence} (fuzzy) approximation.
    Existing methods correspond to specific combinations of exact or approximate operations, but the framework also allows for new TT-based \apricot strategies yet to be explored.
    }
    \label{fig:pipeline}
\end{figure}

In this paper, we provide a unified view over many NeSy pipelines and we identify three stages 
at which approximations can enter (\cref{fig:pipeline}):
\textbf{(I)}~compiling each factor exactly or with a relaxed circuit,
\textbf{(II)}~multiplying factors exactly or via a recompression that
controls size growth, 
and \textbf{(III)}~combining remaining factors under an independence assumption. 
Many existing methods fit in this categorization, sometimes mixing several approximations. 
\emph{Independence} (fuzzy) approximations are by far the most common~\cite{badreddine2020ltn,vankrieken2022fuzzy,ahmed2023semstrenghtening} and are well known to support effective learning~\cite{vankrieken2022fuzzy,maene-hardness_2024}, yet their inference fidelity is not well-studied.
Stages (I) and (II) are far less explored.
Drawing on the close connection between tensor networks and
circuits~\cite{loconte2024relationship}, we instantiate stages (I) and (II) using tensor train (TT) decompositions.
TTs have been used in NeSy to approximate and compress isolated factors~\cite{choi_ctsketch_2025} but not in a full compilation pipeline.
We show how approximate tensor trains can be used for inference and compilation, deriving novel \textbf{\apricot} pipelines.

\textbf{Contributions.}
\textbf{(i)}~We unify existing bottom-up compilation
methods under a three-stage approximate-compilation pipeline and
analyze how error and cost compose across strategies.
\textbf{(ii)}~Within this framework, we derive new approximation strategies based on TN recompressions, previously unexplored in NeSy or compilation.
\textbf{(iii)}~We map the compilation design space on Sudoku and evaluate weighted inference and learning on arithmetic and scene understanding tasks. 
We show that \apricot pipelines scale beyond exact compilation, produce WMC estimates orders of magnitude more accurate than prior methods by preserving dependencies, which also translates to better learning in programs where factors are structurally tightly connected.
Code and instructions for reproducing all experiments are available at \href{https://github.com/sbadredd/apricot}{github.com/sbadredd/apricot}.

\section{Probabilistic NeSy Models}
\label{sec:background}

A probabilistic NeSy model combines (i) a parameterized (neural) model $p_\btheta(\bc \mid \bx)$ which maps the input $\bx$ to a vector of discrete, symbolic \emph{concepts} $\bc = [c_i]_{i = 1, \dots, d}$, $c_i \in \cC_i$, with (ii) a symbolic \emph{constraint} $\phi : \cC_1 \times \dots \times \cC_d \to \{0,1\}$ that assesses the validity of the predicted concepts.
In Sudoku, for example, $\bx$ is the image of the board, $\bc$ are the predicted digits in each cell, and $\phi$ checks that the predicted digits satisfy the Sudoku rules.
This involves computing $p_\btheta(\phi | \bx)$ which maps the input $\bx$ to the probability of satisfying the constraint $\phi$. This is known as 
a conditional \emph{weighted model count} (WMC) problem, defined as~\footnote{The problem can also be stated with $\phi(\bc,\by)$ being a program that maps concepts $\bc$ to a structured output $\by$, in which case the WMC derives $p_\btheta(\by|\bx)$. See the \emph{Sudoku completion} example in \cref{sec:experiments}.}
\begin{equation}
    \label{eq:wmc}
    p_\btheta(\phi \mid \bx) := \sum\nolimits_{\bc \in \cC} p_\btheta(\bc \mid \bx)\  \phi(\bc).
\end{equation}

The problem can also be stated with $\phi(\bc,\by)$ being a program that maps concepts $\bc$ to a supervised structured output $\by$, in which case the WMC derives $p_\btheta(\by\mid\bx)$ (sometimes up to a normalization constant, see \cref{app:conditional_wmc} for details).

Often, $\phi$ factorizes into a conjunction of subformulas $\phi_1(\bc_1) \land \dots \land \phi_N(\bc_N)$, where each factor $\phi_j$ has scope $\bc_j \subseteq \bc$. 
For example, the $9\times 9$ Sudoku rules decompose into $N=810$ factors that check pairwise constraints $c_{i} \neq c_{j}$ for cells $i,j$ in the same row, column, or box.
Programs may additionally introduce \emph{auxiliary symbolic variables}, unweighted and unsupervised, to derive the factorization.
For example, the arithmetic rules of addition, often explored in NeSy~\cite{daniele_DSL_2023,maene_softunification_2023}, factorize using auxiliary carries between digits.
This is further discussed in \cref{app:conditional_wmc}.

Computing the WMC is \#P-hard in general~\cite{valiant1979enumeration}, and this is also due to the large combinatoric space of NeSy problems.
E.g., in a $9\times 9$ Sudoku there are
$9^{81}$ possible configurations, out of which \textit{only} $6.67 \times 10^{21}$ are valid boards~\cite{Felgenhauer2006MathematicsOS}.
To solve this issue, probabilistic NeSy methods must use compilation techniques and/or relaxation strategies to make the WMC tractable.

\subsection{Exact circuit compilation}
\label{sec:exact_circuit_compilation}
A prominent approach to computing the WMC tractably is to represent both $\phi$ and $p_\btheta(\bc|\bx)$ as circuits and combine them \citep{kisa2014probabilistic,manhaeve2018deepproblog,xu2018semantic,ahmed2022semantic}.
A circuit is a directed acyclic computational graph of \emph{weighted sum} and \emph{product operations} over tractable input functions \citep{choi2020pc,vergari2021compositional}.
We focus on \emph{smooth} and \emph{decomposable} circuits, which support marginalization in time linear in their size.
For tractable multiplication, circuits additionally need compatible decomposition structures: a circuit is \emph{structured-decomposable} when its product units respect a common hierarchical partition of the variables (called a vtree), and two such circuits are \emph{compatible} when these decompositions agree on shared variables \citep{vergari2021compositional}.
Compatible circuits can be multiplied in polynomial time while preserving smoothness and decomposability.
We defer formal definitions to \cref{app:circuits}.

\hlkc{\textbf{Knowledge compilation}} (KC) transforms a logical constraint $\phi$ into a logical circuit, typically expressed using AND/OR gates \citep{darwiche2002knowledge}.
For deterministic circuits, whose OR children have disjoint supports, AND/OR gates can be interpreted as product/sum operations without changing the represented Boolean function \citep{kimmig2017algebraic}.
A deterministic compiled circuit for $\phi$ can be multiplied tractably with a compatible probabilistic circuit $p_\btheta(\bc|\bx)$.

Sentential decision diagrams (SDDs) \citep{darwiche2011sdd} are a popular class of deterministic, structured-decomposable logical circuits.
They can also be constructed bottom-up by \hlexact{\textbf{multiplying}} circuit representations of factors $\phi_j$.
Let $\hlkc{\KC}(\phi_j)$ denote the circuit obtained by compiling factor $\phi_j$.
We express bottom-up compilation using $\hlexact{\MUL}$:
\begin{equation}
\sum\nolimits_{\bc \in \cC}
p_\btheta(\bc|\bx)
\hlexact{\MUL}
\big(
\hlkc{\KC}(\phi_1), \dots, \hlkc{\KC}(\phi_N)
\big)(\bc).
\end{equation}
This is the classical recipe behind many approaches such as semantic loss \citep{xu2018semantic} or DeepProbLog \citep{manhaeve2018deepproblog}.

The main limitation is constructing a sufficiently compact circuit for $\phi$ with the required structural properties.
KC is at least NP-hard \citep{darwiche2002knowledge,chavira2008wmc}, and, to the best of our knowledge, existing compilers cannot produce a polynomial-size circuit in reasonable time for a $9\times9$ Sudoku instance, even when relaxing structured decomposability \citep{darwiche2001decomposable,lagniez_2017_d4}.~\footnote{Current boolean KC software also commonly requires categorical variables to be propositionalized.
This requires introducing mutual exclusivity constraints which further increases the size of the formula and complicates compilation. We discuss the resulting encoding overhead in \cref{app:circuits}.}
Therefore, a common practice is to consider \textit{approximations}.
We focus on approximations which (i) apply to bottom-up circuit compilation, (ii) are differentiable. In \cref{app:related_work}, we discuss other approaches to WMC in NeSy, such as amortized or sampling-based methods, and how they relate to our work.

\subsection{Approximations}

\paragraph{\hlindepb{Independence} approximations}
A common approximation when multiplying two circuits is to assume they are independent, even if they are defined over the same set of variables.
If we assume that, the WMC of their conjunction factorizes into the product of their WMCs.
More formally,
\begin{equation}
\label{eq:indep}
    \sum\nolimits_{\bc \in \cC} p_\btheta(\bc|\bx) \hlexact{\MUL}(\hlkc{\KC}(\phi_1), \dots, \hlkc{\KC}(\phi_N))(\bc) \stackrel{\indep}{\approx} \hlindep{\prod\nolimits_j} \left(\sum\nolimits_{\bc_j \in \cC_j} p_\btheta(\bc_j|\bx) \hlkc{\KC}(\phi_j)(\bc_j)\right).
\end{equation}

Often, $|\cC_j| \ll |\cC|$, which makes the factor WMCs much easier to compute.
\emph{Fuzzy} NeSy approaches with product semantics~\cite{badreddine2020ltn,donadello2017ltnsii,vankrieken2022fuzzy} can be understood as probabilistic approaches relaxing \emph{all} dependencies~\cite{vankrieken_2019_fuzzyvsprod}.~\footnote{In the rest of the paper, we sometimes refer to the independence approximation as ``fuzzy'', though it is important to note that not all fuzzy approaches use product semantics understood as an independence relaxation of probabilistic NeSy.}
The approximation is only correct when the factors $\phi_j$ are independent, i.e., when the scopes $\bc_j$ are pairwise disjoint. 
In general, this is not the case and the approximation can be arbitrarily bad, as we will show in our experiments.
Still, in the context of differentiable learning, fuzzy relaxations are known to be effective in practice~\cite{vankrieken2022fuzzy}. 

Fuzzy relaxations can be used in combination with exact circuit compilation.
\SemStrengthening~\cite{ahmed2023semstrenghtening} compiles SDD circuits for factor groups under a certain budget, prioritizing groups of factors with high mutual information, and applies the independence approximation between groups.
SDDs are particularly suitable for this approach as they can iteratively multiply and analyze factors, making informed compilation or stopping decisions within the budget.

\paragraph{\hlfactorb{Factor} approximations}
Instead of approximating the marginalization step, one can approximate the factors themselves.
This requires approximate compilation ($\hlfactor{\operatorname{AC}}$) step, i.e.,
\begin{equation}
    \MUL(\hlkc{\KC}(\phi_1), \dots, \hlkc{\KC}(\phi_N)) \stackrel{\fac}{\approx} \MUL(\hlfactor{\operatorname{AC}}(\phi_1), \dots, \hlfactor{\operatorname{AC}}(\phi_N))
\end{equation}
where $\hlfactor{\operatorname{AC}}$ compiles an \emph{approximated circuit} in the real semiring for factor $\phi_j$, i.e., a circuit that does not encode a Boolean function but closely approximates it up to an $\epsilon$-error.
Crucially, the circuit retains all the structural properties needed to tractably compute $\MUL$.
In \cref{sec:tensor_networks}, we discuss how each factor can be approximated with \emph{tensor trains} (TTs)~\cite{oseledets2011tensor}, a special family of tensorized circuits \citep{loconte2024relationship} for which there is a rich literature on function approximation \citep{orus2013practical}.
Such a TT-approximated factor $\hlfactor{\operatorname{AC}}(\phi_j)$ can be a representation that is significantly smaller than the exact circuit $\hlkc{\KC}(\phi_j)$, given that it can introduce negative parameters \citep{loconte2024subtractive}.
Additionally, TTs inherently handle categorical variables and do not need Boolean encoding. 

\CTSketch~\cite{choi_ctsketch_2025} is a recent approach that approximates factors $\phi_j$ with low-rank TTs.
However, \CTSketch only considers programs that follow a certain tree-like independence structure, where any variable $c_i$ can appear in at most one factor.
This applies to simple problems such as \MNISTAdd~\cite{manhaeve2018deepproblog}, but most programs, like Sudoku boards, have overlapping factors and violate this assumption.
Therefore, \CTSketch mixes factor approximations with fuzzy independence approximations $\hlindep{\prod_j} (\sum_{\bc_j} p_\btheta(\bc_j | \bx) \hlfactor{\operatorname{AC}}(\phi_j)(\bc_j))$.
\cref{app:ctsketch} details this.
In \cref{sec:tensor_networks}, we show how to exactly multiply TT-approximated factors under more complex dependence structures.

\paragraph{\hlprodb{Product} approximations}
Finally, we consider approximating the multiplication step directly, without going through intermediate approximations of the factors themselves, i.e.,
\begin{equation}
    \hlexact{\MUL}(\hlkc{\KC}(\phi_1), \dots, \hlkc{\KC}(\phi_N)) \stackrel{\prodap}{\approx} \hlprod{\RMUL}(\hlkc{\KC}(\phi_1), \dots, \hlkc{\KC}(\phi_N)).
\end{equation}
Drawing on TN theory, we show that it is possible to derive algebraically-principled approximations which compress the product of circuits into smaller circuits.
We denote such operations $\hlprod{\RMUL}$ for \emph{rounded multiplications}, inspired by the \TTround~\cite{oseledets2011tensor} algorithm discussed in \cref{sec:tt_approx_mult}.
This direction is, to the best of our knowledge, unexplored in the NeSy or compilation literature, and can prove useful to approximate many operations on circuits.

\section{Approximate Tensor Trains...}
\label{sec:tensor_networks}

The tensor perspective applies to WMC by representing the constraint $\phi$ as a Boolean-valued tensor $\bcX \in \{0,1\}^{n_1 \times \dots \times n_d}$, where $n_i = |\cC_i|$ and $\bcX(\bc) := \phi(\bc)$.
This is an order-$d$ tensor: each mode corresponds to one variable $c_i$, and its size $n_i$ is the number of values that variable can take. Thus, each tensor entry records whether a particular assignment satisfies the constraint.
The dense representation contains $\prod_{i=1}^d n_i$ entries,
which quickly becomes infeasible to materialize as the number of variables grows.
For example, the full $9\times9$ Sudoku constraint has $d=81$
variables, each with $9$ possible values, giving an order-$81$
tensor with $9^{81}$ entries.

Tensor trains (TTs)~\cite{oseledets2011tensor,perez2006mps}
represent such tensors through a sequence of smaller tensors.
They are a special case of tensor networks and can also be viewed
as structured-decomposable circuits~\cite{loconte2024relationship}.
A TT represents an approximation $\approxhat{\bcX}$ of an
order-$d$ tensor $\bcX$ using one \emph{core} per mode:
\[
    \bcX(c_1,\dots,c_d)
    \approx
    \approxhat{\bcX}(c_1,\dots,c_d)
    =
    \bcT_{\approxhat{\bcX},1}(c_1)
    \cdots
    \bcT_{\approxhat{\bcX},d}(c_d).
\]
Each core $\bcT_{\approxhat{\bcX},k}$ is an order-$3$ tensor
of shape $r_{k-1}\times n_k\times r_k$.
Fixing the variable value $c_k$ selects a matrix slice
$\bcT_{\approxhat{\bcX},k}(c_k)
:=\bcT_{\approxhat{\bcX},k}(:,c_k,:)$.
The dimensions $r_k$ are the \emph{TT ranks}, with
$r_0=r_d=1$ so that multiplying all selected matrices returns
a scalar.
Writing $n=\max_k n_k$ and $r=\max_k r_k$, TT storage is
$\cO(dnr^2)$, compared with up to $\cO(n^d)$ entries in the
dense tensor.
The representation is therefore compact when the ranks remain small.
We now use this representation to approximate individual constraint factors, multiply the resulting TTs, and recompress their products.

\begin{figure}
    \centering
    \begin{subfigure}[b]{0.32\textwidth}
        \centering
        \includegraphics[width=\textwidth]{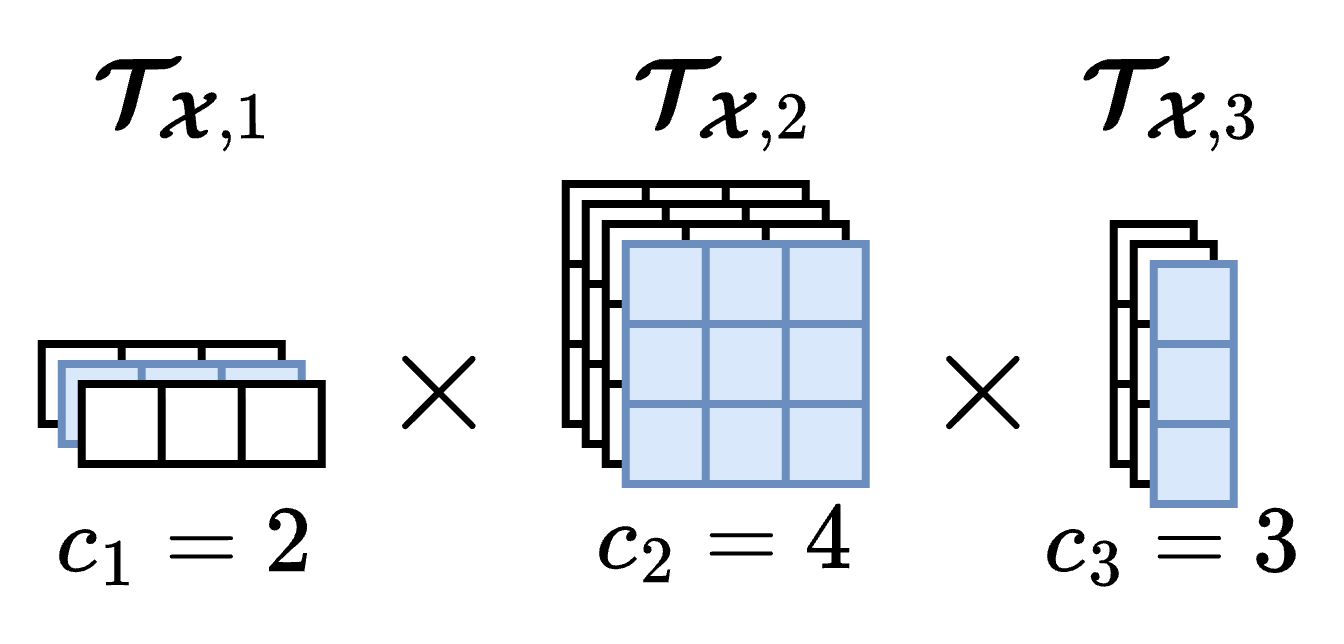}
        \caption{$\hlfactor{\bcX(2,4,3)}$}
        \label{fig:tt-algebra-a}
    \end{subfigure}
    \qquad
    \begin{subfigure}[b]{0.24\textwidth}
        \centering
        \includegraphics[width=\textwidth]{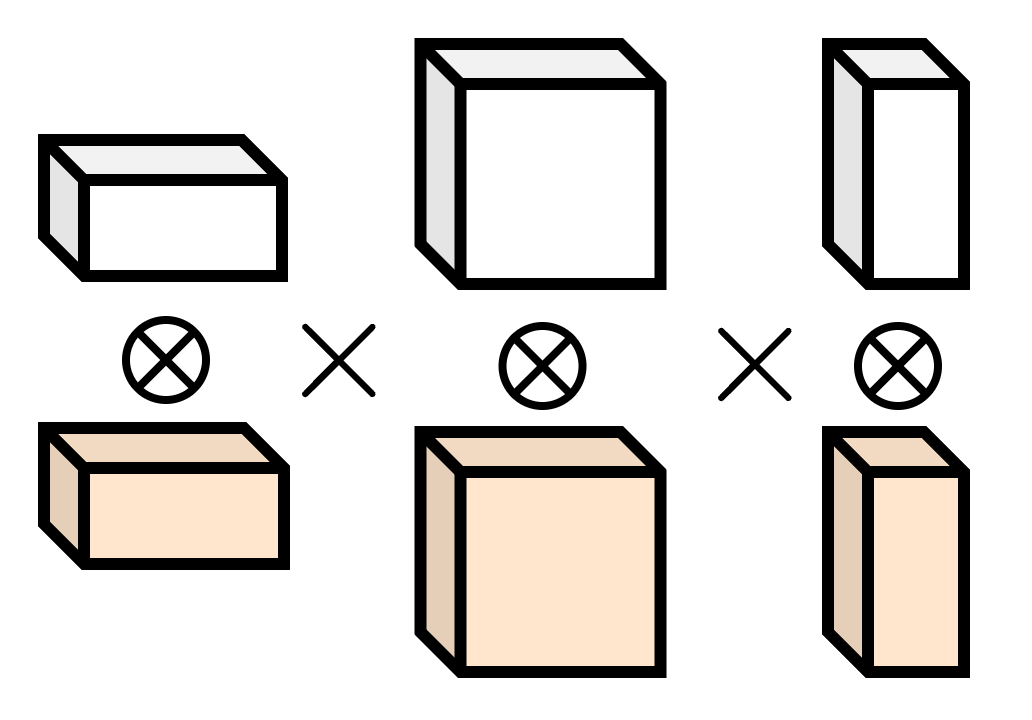}
        \caption{$\bcX_j \hlexact{* \bcX_l}$}
        \label{fig:tt-algebra-b}
    \end{subfigure}
    \qquad
    \begin{subfigure}[b]{0.32\textwidth}
        \centering
        \includegraphics[width=\textwidth]{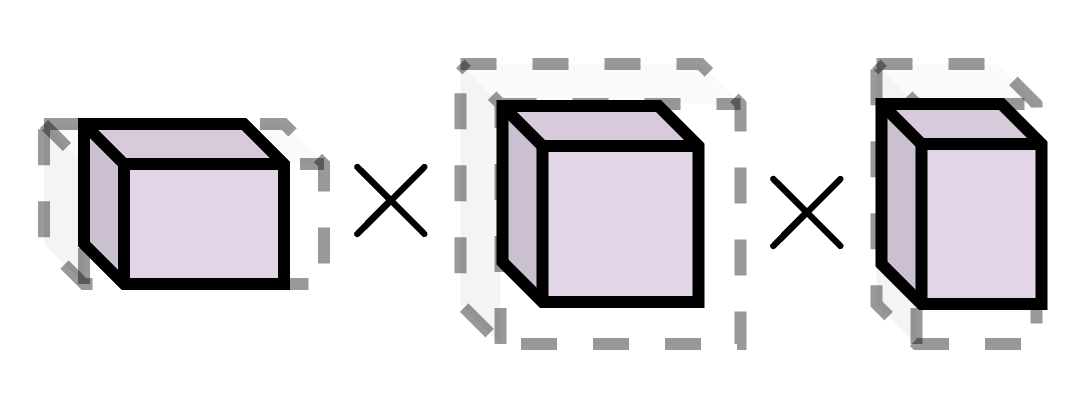}
        \caption{$\hlprod{\TTround}$}
        \label{fig:tt-algebra-c}
    \end{subfigure}
    \caption{Closed algebra over TTs.
    TTs (a) can \hlfactorb{approximate discrete functions} under an error criterion and (b) are closed under several \hlexactb{exact operations} including multiplications.
    As multiplications involve Kronecker products $\otimes$ of the cores resulting in an exponential increase in rank, algebraic recompression strategies (c) can \hlprodb{approximate the result of the operations}.}
    \label{fig:tt_algebra}
\end{figure}

\subsection{\hlfactorb{TT Factor} Approximations}
\label{sec:tt_factor_approx}
A factor $\phi_j$ with scope $\bc_j$ of size $d_j$ corresponds to an order-$d_j$ tensor $\bcX_j$.
For example, each pairwise Sudoku constraint $c_i\neq c_l$ is an order-$2$ tensor: a $9 \times 9$ matrix with zeros on the diagonal and ones elsewhere.
While the full constraint tensor is generally too large to materialize, these factor tensors are often small enough to construct explicitly and approximate using \TTSVD.

We approximate each dense factor tensor using \TTSVD~\cite{oseledets2011tensor}, as in
\CTSketch~\cite{choi_ctsketch_2025}. The algorithm constructs a TT $\approxhat{\bcX}_j$ under either a prescribed Frobenius error tolerance $\|\bcX_j-\approxhat{\bcX}_j\|_F\leq\epsilon_j$ or a maximum TT rank $R$.
For prescribed ranks, \TTSVD is quasi-optimal: its error is within a factor $\sqrt{d_j-1}$ of the best approximation with those ranks.
Importantly, \TTSVD requires its dense input to be materialized, which motivates applying it to the small factor tensors rather than to the full constraint.%
The resulting TTs must then be combined (i.e., multiplied) while accounting for dependencies between their overlapping scopes.

\subsection{\hlexactb{Exact multiplication } of TTs}
\label{sec:tt_exact_mult}
TTs are closed under element-wise (Hadamard) multiplication, which allows us to compile TT-approximated factors without introducing new approximations.
The operation requires matching scopes and mode orderings.
We fix a global variable ordering and arrange each factor's modes accordingly. 
Before multiplying factors with different scopes, we \emph{broadcast} their TTs to the union of their scopes by inserting identity cores for missing variables.
The construction and a heuristic for choosing the global ordering %
are detailed in \cref{app:tt_broadcasting}.

Consider two TTs $\bcX_j$ and $\bcX_l$ after broadcasting to
a common scope of $q$ variables, locally indexed
$c_1,\dots,c_q$.
Their product $\bcX^*=\bcX_j*\bcX_l$ has cores
\begin{equation}
    \label{eq:tt-mult}
    \bcT_{\bcX^*,k}(c_k)
    =
    \bcT_{\bcX_j,k}(c_k)
    \otimes
    \bcT_{\bcX_l,k}(c_k),
    \qquad k=1,\dots,q,
\end{equation}
where $\otimes$ denotes the matrix Kronecker product.
This is a special case of the $\hlexact{\MUL}$ operation
for compatible structured-decomposable circuits which maintains a TT structure.
Because of the Kronecker, the product ranks are $r_k^*=r_k^j r_k^l$.
Thus, two TTs with ranks bounded by $r$ can produce
a TT with ranks up to $r^2$ and storage $\cO(qnr^4)$.
Repeatedly multiplying $m$ such TTs can require
$\cO(qnr^{2m})$ storage in the uncompressed representation.
This matches the worst-case exponential complexity of circuit multiplication~\cite{vergari2021compositional}.
Exact multiplication therefore preserves dependencies
between factors, but its rank growth motivates
recompression.

\subsection{\hlprodb{Approximate multiplications of} TTs}
\label{sec:tt_approx_mult}
To control rank growth, we approximate the product $\bcX^*$ by a TT with a target maximum rank $R$.
Unlike factor approximation with \TTSVD, this \emph{product recompression} operates directly on the input TTs, without materializing the dense product.
Depending on the method, the recompression can alternatively be controlled by a truncation tolerance $\epsilon$.
We consider four strategies, distinguished by their compression mechanism (SVD or interpolation) and their use of deterministic or randomized compression directions.
Algorithmic details are deferred to \cref{app:tt_mult_algos}.

\TTround~\cite{oseledets2011tensor} \emph{(SVD, deterministic/site-randomized)} uses successive SVD-based truncations to reduce TT ranks.
For Hadamard products, we exploit the Kronecker structure to interleave products and compressions without materializing the Kronecker product cores.
We consider both deterministic and randomized per-site SVDs~\cite{halko2010randomsvd}.

\HaTT~\cite{sun2025hatt} \emph{(SVD, randomized)} uses a random TT sketch of the
product to identify compression subspaces, then applies SVDs to small sketched matrices. Its randomization acts at the TT level, rather than independently within each local SVD.

\TCI~\cite{NunezFernandes2025TCI,oseledets_tt-cross_2010} \emph{(interpolation, deterministic)} constructs a TT approximation from selected entries of the product, choosing representative rows and columns through maxvol-based cross interpolation~\cite{goreinov2010maxvol}.

\RSI~\cite{meng2026RSI} \emph{(interpolation, randomized)} combines random sketching with interpolative decompositions. It replaces \TCI's maxvol-based initialization with sketched environments built from the input TTs. We extend its original single left-to-right sweep to alternating left-to-right and right-to-left sweeps, using the connection to \TCI (details in \cref{app:rsi}).

\subsection{From tensor error to WMC error}
The Frobenius guarantees provided by \TTSVD and SVD-based recompression translate to a bound on the final inference error.
Let $\approxhat{\bcX}$ approximate the full constraint tensor $\bcX$, with $\|\approxhat{\bcX}-\bcX\|_F\leq\epsilon$, and let $\bcP(\bc)=p_\btheta(\bc\mid\bx)$.
Since the WMC $p_\btheta(\phi|\bx)$ in \cref{eq:wmc} is the Frobenius inner product of these tensors, Cauchy-Schwarz gives
\begin{equation}
    \label{eq:wmc_frobenius_bound}
    \left|\wmchat-\wmc\right|
    \leq \epsilon\,\|\bcP\|_F
    \leq \epsilon.
\end{equation}
The last inequality follows because $\bcP$ is a normalized probability tensor.
More details are given in \cref{app:error_bounds}, including how the factor errors $\epsilon$ propagate under multiplications.

\section{\dots for Inference and Compilation}
\label{sec:pipeline}
We now combine these ingredients into the TT-based framework \apricot:
\textbf{AP}proximate \textbf{R}epresentations for
\textbf{I}nference and \textbf{CO}mpilation using
\textbf{T}ensor trains.
We describe instantiations of this framework within a unified bottom-up pipeline that also accommodates exact circuit compilation and some pre-existing approximation strategies (\cref{fig:pipeline}). 
Given a constraint $\phi=\phi_1\land\dots\land\phi_N$, the pipeline has three stages: compiling factors, multiplying their circuit representations, and evaluating the remaining circuits.
The choices at each stage decide whether \hlfactorb{factor}, \hlprodb{product}, and/or \hlindepb{independence} approximations are involved.

\paragraph{Step I: \hlfactorb{Factor compilation}}
Each factor is represented by a circuit. Exact compilation constructs $\hlkc{\KC}(\phi_j)$, for which we use SDDs~\cite{darwiche2011sdd}.
Alternatively, \apricot applies \hlfactorb{\TTSVD} to the dense factor tensor $\bcX_j$, with an initial rank budget $\hlfactor{R_{\text{init}}}$ or a Frobenius error tolerance (\cref{sec:tt_factor_approx}).
Before TT decomposition, small factors may optionally be grouped and \hlexactb{pre-multiplied} exactly in dense tensor space.
This captures dependencies within each group before introducing a factor approximation, as long as the grouped tensor fits in memory.
Let $M$ denote the number of circuits resulting from this stage. Without grouping, $M=N$.
All circuits use compatible decomposition structures (in the case of TTs, variable orderings) so that they can subsequently be multiplied.

\paragraph{Step II: \hlprodb{Multiplication} ($M\to K$)}
The $M$ circuits are reduced to $K$ circuits through pairwise multiplications, where $1\leq K\leq M$.
\hlexactb{Exact multiplication} uses SDD Apply for SDDs or the TT product of \cref{sec:tt_exact_mult}.
For \apricot, the propagation of existing factor errors is analyzed in \cref{app:error_bounds}.
To control the growth of intermediate TTs, \apricot can instead use \hlprodb{recompressed multiplication}, with any of the strategies introduced in \cref{sec:tt_approx_mult}.
Recompression introduces additional approximation error in exchange for smaller representations.
The multiplication schedule, recompression strategy, and stopping criterion are all parameters of the pipeline. 
For example, $K$ may be prescribed or determined by the available resources and approximation conditions.
Our experimental choices are specified in \cref{sec:experiments}.

\paragraph{Step III: Final \hlindepb{independent reduction} ($K\to1$)}
Let $\approxhat{\phi}_1,\dots,\approxhat{\phi}_K$ be the remaining circuits, with scopes $\bc_1,\dots,\bc_K$.
We compute their individual WMCs and combine them as
$$
\widehat{p}_{\btheta}(\phi\mid\bx)
    =
    \hlindep{\prod\nolimits_{k=1}^{K}}
    \left(
        \sum\nolimits_{\bc_k\in\cC_k}
        p_{\btheta}(\bc_k\mid\bx)\,
        \approxhat{\phi}_k(\bc_k)
    \right),
$$
where $\cC_k$ is the assignment space of the $k$-th residual scope.
For $K>1$, this applies the \hlindepb{independence} approximation between remaining factors (\cref{eq:indep}).
The introduced error is difficult to predict and depends on the specific factors remaining after Step II (\cref{app:indep_categorical_binary}).

\Cref{fig:pipeline} maps existing approaches and new \apricot variants as different choices in each step of the pipeline.
Exact compilation and multiplication to $K=1$ is exact inference, while stopping earlier gives the grouped independence strategy used, for example, by \SemStrengthening.
TT factor approximation followed directly by independent reduction gives the \CTSketch-style approximation for overlapping factors, a special case of \apricot.
Variants can additionally combine them: \apricot A combines TT factors through exact multiplications before an independence reduction, \apricot B additionally uses product recompression up to a budget, while \apricot C avoids the final independence approximation.
Notice that each row in \cref{fig:pipeline} identifies a family rather than a fixed configuration: ranks, multiplication schedules, and reduction targets can vary across tasks.
In particular, we detail different multiplication schedules we evaluate in \cref{app:partitioning_mul}. 

\section{Experiments}
\label{sec:experiments}
We first explore the compilation design space on Sudoku, a popular NeSy benchmark~\cite{augustine_2022_visudopc,wang_2019_satnet} with a complex dependence structure, and evaluate in-depth where to introduce approximation during compilation.
We then compare representative pipelines on arithmetic and scene-understanding tasks of increasing difficulty.
Finally, we address whether accurately compiling dependencies improves neurosymbolic learning across these settings.

\paragraph{Tasks} 
For the compilation ablation, we use Sudoku boards of size $4\times4$, $6\times6$, and $9\times9$, where known model counts derived analytically~\cite{Felgenhauer2006MathematicsOS} allow us to evaluate unweighted approximation quality.
For learning, we use two Visual Sudoku tasks. \textbf{ViSudo-Classification} predicts whether a board of MNIST images satisfies the Sudoku rules, using the dataset from \citet{augustine_2022_visudopc}. \textbf{ViSudo-Completion} predicts missing cells from the observed digit images, using the SatNet dataset~\cite{wang_2019_satnet} with the data-leak correction of \citet{chang2023assessingsatnet}.
We additionally evaluate four tasks for both weighted inference and learning.
\textbf{MNISTAdd}~\cite{manhaeve2018deepproblog} and \textbf{MNISTMul} predict the sum or product of two $L$-digit numbers represented by MNIST images, with supervision only on the result.
\textbf{CLEVR-Diff} predicts whether all objects in a CLEVR scene~\cite{johnson2017clevr}
have different attribute tuples (color, shape, \dots), while \textbf{CLEVR-CountDiff} predicts the number of distinct attribute tuples in the scene.
We construct the CLEVR tasks reusing scenes from \citet{stammer_right_2021}.

These tasks differ in how their programs factorize. 
In particular, the arithmetic and distinct-counting programs introduce \emph{auxiliary symbolic variables} such as carries and running counts to express the constraint with smaller factors (see \cref{app:programs} for details). 
In comparison, our Sudoku and CLEVR-Diff factorizations directly compare perceptual concepts and do not introduce such intermediate symbols. 
This will prove meaningful when analyzing the NeSy learning results.

\paragraph{Baselines}
We use the family names introduced in \cref{fig:pipeline}.
The Sudoku compilation ablation explores many configurations within these families, while the learning experiments evaluate only selected representatives.
On the arithmetic and CLEVR tasks, we compare three strategies: full independence reduction (LTN/DFL-style), full exact compilation (Semantic Loss/DeepProbLog-style), and full TT compilation allowing product recompression (\apricot C).
As explained in \cref{sec:pipeline}, the names identify families; the exact configurations (ranks, multiplication schedules) depend on the specific task programs and are detailed in \cref{app:traineval_details}.

We implement all TT algorithms in PyTorch~\cite{paszke_2019_pytorch}, use PySDD~\cite{meert2017pysdd} for SDD compilation, and KLay~\cite{maene2025klay} for GPU inference on SDDs. 
Results are averaged over ten runs with errors bars for standard deviation.
Hardware and training protocols are given in \cref{app:traineval_details}.

\subsection{Sudoku compilation ablation}
\label{sec:sudoku_ablation}
\begin{figure}
    \centering
    \begin{subfigure}[b]{0.48\textwidth}
        \centering
        \includegraphics[width=\textwidth]{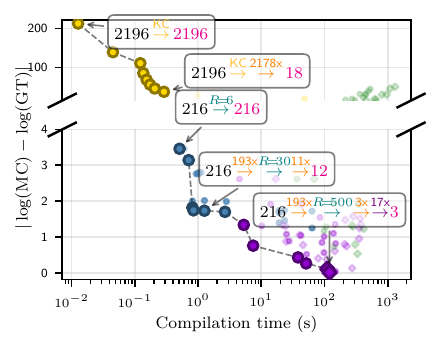}
    \end{subfigure}
    \hfill
    \begin{subfigure}[b]{0.470\textwidth}
        \centering
        \includegraphics[width=\textwidth]{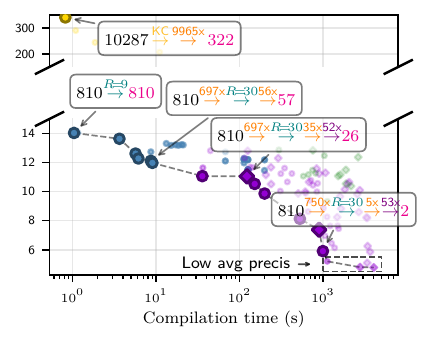}
    \end{subfigure}
    \caption{Pareto front of absolute log inference error vs. compilation time for Sudoku 6x6 (left) and 9x9 (right).
    SDD pipelines are in yellow \marker{yellow}{circle}. \apricot pipelines are in blue \marker{blue}{circle} when they do not use recompressions, in violet when they use SVD-based recompressions (\TTround \marker{violet}{circle} and \HaTT \marker{violet}{diamond}) and in green for interpolation-based strategies (\TCI \marker{ForestGreen}{circle} and \RSI \marker{ForestGreen}{diamond}).
    }
    \label{fig:pareto-inference}
\end{figure}

\Cref{fig:pareto-inference} compares compilation time and model-count error on the larger boards.
We vary the number and type of the multiplications (exact or recompressed) as well as the rank for the initial TT-SVD.
Each point in the plot is a candidate compilation pipeline; details on our grid search and on the tags we use to identify points are in \cref{app:sudoku_ablation}.
Below, we highlight trends in the results.

We measure absolute errors on a logarithmic scale $|\log_{10} \mathrm{MC}/\mathrm{MC}_{\text{ref}}|$, with zero indicating an exact count.
The model count estimates are obtained by calculating the WMC under a uniform distribution, which gives the probability of an assignment being a model, multiplied by the size of the assignment space.
The ground-truth counts are $288$, $28{,}200{,}960$, and approximately $6.67\times10^{21}$ for the three board sizes.
$4\times4$ serves as a sanity check, with both exact SDD compilation and suitable TT configurations recovering its count to numerical precision (\cref{app:pareto_n4}).
$6 \times 6$ is already challenging: compilation is only possible with the best \emph{non}-structured-decomposable compilers like d4~\cite{lagniez_2017_d4}, not with SDDs. 
To the best of our knowledge, $9 \times 9$ is unsolved by any existing compilers. %

\paragraph{On \hlindepb{independence} approximations} As expected from our error analysis and discussion on fuzzy bounds (\cref{app:error_bounds}), the \emph{fuzzy approximations of the WMC have unpredictable behaviors} which vary with the encoding used. 
    Even at matched $K$, SDD (yellow) and \apricot behave very differently. In practice, the MC estimate of SDD+Fuzzy is approximately zero, making it hardly usable.

\paragraph{On \hlexactb{exact} multiplications} Combining factors before independence reduction always \emph{significantly improves fidelity}. 
However, both SDD size and TT ranks limit how far exact multiplication scales. 
\hlfactorb{Factor} approximations can compress initial factors, but their products still grow exponentially.

\paragraph{On \hlprodb{approximate} multiplications}
Recompression allows further multiplications and provides the strongest accuracies.
The \emph{SVD-based strategies are more reliable} than the interpolation-based alternatives, with no significant difference between randomization strategies.
In both cases, however, attempting too many multiplications under a limited rank budget can also degrade the representation because of expressivity limitations.
To distinguish count accuracy from constraint fidelity, we additionally evaluate how well each circuit separates satisfying assignments from nonsatisfying assignments.
We measure this using average precision on 1000 model and 1000 non-model sudoku samples, evaluating the product of the current factors indexed on each sample.
On $9 \times 9$, we find points (marked ``low AP'') which, after much recompression, achieve good aggregate error but miss the actual models.
This highlights the importance of halting compilation based on the accumulated Frobenius error bound (\cref{app:error_bounds}) once it exceeds a certain tolerance.

\subsection{Weighted inference}
\label{sec:weighted_inference}
\begin{figure}
    \centering
    \includegraphics[width=\textwidth]{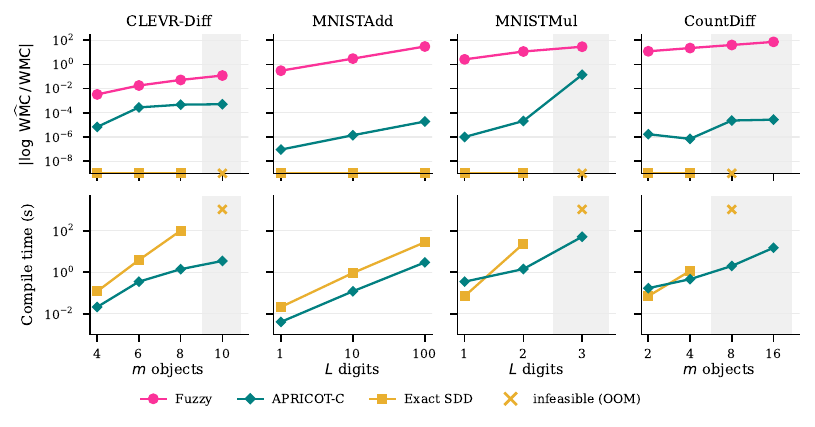}
    \caption{Median absolute log WMC error and compilation time across tasks and problem sizes.
    Exact SDD compilation becomes infeasible on harder problems.
    Full TT reduction (\apricot C) can scale further while producing estimates orders of magnitude more accurate than fuzzy.}
    \label{fig:weighted_inference}
\end{figure}

Next, we evaluate whether the compilation strategies remain effective when querying the circuits with weights.
We compare circuit estimates with exact compilation or task-specific exact references under 64 shared random weightings. 
For each weighting, we compute the absolute base-10 log-ratio between the estimated and reference WMC, and report the median of these 64 errors alongside compilation time (\cref{fig:weighted_inference}).
That is, a reported error of $1$ means that, for a typical sampled weighting, the estimated WMC differs from the reference by a factor of $10$.
Details are specified in \cref{app:wmc_evaluation}.

On MNISTAdd, both SDDs with a suitable vtree and \apricot exploit the carry-chain structure to scale efficiently with the number of digits. On the more challenging CLEVR-Diff, MNISTMul, and CLEVR-CountDiff, \apricot C scales beyond the evaluated SDD compiler.
At the same time, its WMC estimates are substantially more accurate than those obtained by independence reduction.
Independence reduction is inexpensive at the cost of inference fidelity, but TT recompression provides a practical trade-off, retaining these dependencies approximately in a manageable representation.

\subsection{Learning}
\label{sec:learning_results}
Previous works have found that faithful inference is not always a requirement for competitive downstream learning.
Indeed, despite ignoring dependencies, fuzzy systems provide good differentiability in many use cases~\cite{badreddine2020ltn} and can behave surprisingly well as gradient estimators~\cite{maene-hardness_2024}. 
We explore whether this holds across more complex settings.

We measure the concept accuracy as the fraction of individual concepts $\bc$ predicted correctly (digits or object keys) and the output accuracy $\by$ as a per-instance exact accuracy (e.g., all missing cells must be correct for Sudoku completion, all output digits must be correct for arithmetic).
Notice that output accuracy can decline with problem size even when individual concepts and reasoning remain accurate: for two inputs with $L$ digits, independent digit errors at $99\%$ accuracy gives a probability of all inputs being correct $0.99^{2L}$ (about $13\%$ at $L=100$), which is necessary for per-instance output accuracy.
For ViSudo-Classification and both CLEVR tasks, we evaluate concept accuracy up to a permutation given that the exact concepts $\bc$ are not identifiable from the label signal~\cite{marconato2023reasoningshortcuts,marconato2026RS}.
More details are provided in \cref{app:traineval_details}.

\begin{figure}
    \centering
    \includegraphics[width=\textwidth]{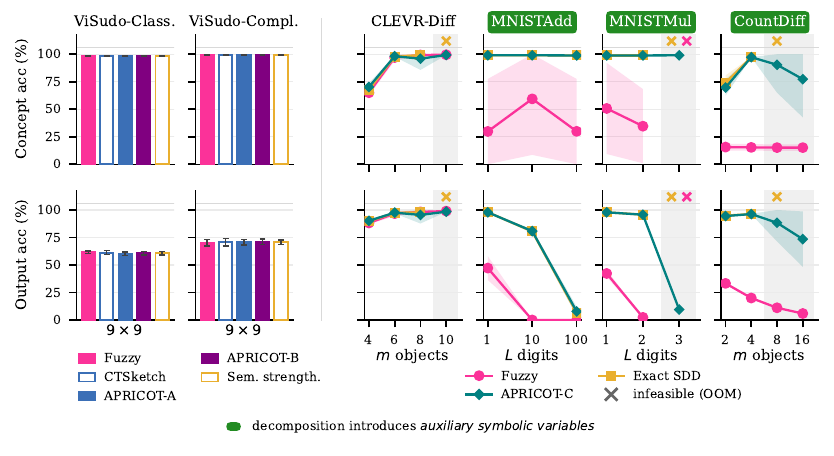}
    \caption{Concept and output accuracy across tasks and problem sizes.
    \apricot matches exact compilation when the latter is feasible, and outperforms fuzzy in larger regimes.}
    \label{fig:learning_results}
\end{figure}

We find a clear contrast depending on the task constraint. 
Independence reduction is competitive on Visual Sudoku and CLEVR-Diff (first three columns in \cref{fig:learning_results}) with all baselines achieving close to $99\%$ concept accuracy, leaving little room for improvement.
However, preserving dependencies improves learning on the arithmetic and CLEVR-CountDiff programs (last three columns).
We hypothesize that this is due to \emph{factorizations introducing \textbf{\textcolor{OliveGreen}{auxiliary symbolic variables}}}.
For example, in MNISTAdd, the constraint (detailed in \cref{app:programs}) has factors at each position $k$ 
$$
A_k + B_k + \textcolor{OliveGreen}{C_{k-1}} = Y_k + 10\textcolor{OliveGreen}{C_k}
$$
where the perception network predicts input digits $A_k, B_k$, derives an output $Y_k$, and introduces carries $\textcolor{OliveGreen}{C_k}$.
The carry $\textcolor{OliveGreen}{C_k}$ is an unweighted auxiliary symbol that denotes the part of the result at position $k$ which exceeds 10 and rolls over position $k+1$.
Its value must agree between the factors at these two positions, and jointly marginalizing them enforces this agreement.
On the other hand, independently marginalizing the factors permits inconsistent carries which weakens the propagation of the supervised signal from $Y_k$ to the perceived $A_k, B_k$.
This matters for learning concepts and inferring outputs in more complicated settings (arithmetic as $N$ grows, or CountDiff for any $m$).

In such settings, exact compilation and \apricot C preserves agreements between factor computations and vastly outperforms the fuzzy baseline.
Notice that this gap is not domain dependent as exemplified by the contrast between CLEVR-Diff and CLEVR-CountDiff, highlighting the importance of the constraint structure.

Larger error bars sometimes reflect optimization failures. 
For example, on CountDiff with $m=8$ and $m=16$, respectively one and three of ten \apricot runs fail to converge, or on arithmetic nearly half of the fuzzy seeds.
Task-specific tuning may reduce these failures, but the aggregate results support the trends discussed above.
We also note that on MNISTMul, fuzzy uses a different factorization because the finer encoding used by SDD and \apricot would give vanishing gradients (details  in \cref{app:constraints}). 
Its dense initial factors are thus larger and exceed GPU memory at $L=3$.

\section{Conclusions}
We unify existing bottom-up compilation methods for WMC under a modular pipeline (\cref{fig:pipeline}) and instantiate previously unexplored configurations via tensor-train decompositions (\apricot).
\apricot combines TT factor representations with exact and recompressed multiplication, improving inference fidelity while extending the scale of feasible compilation.
Our learning experiments show that independence reduction can be effective, but that preserving dependencies matters when local computations must relate shared auxiliary variables.

\paragraph{Limitations} 
Most TT decompositions and recompressions -- particularly \TCI and \RSI -- are not specifically tailored to sparse Boolean tensors.
Future work should develop methods specific to this regime.
Our results also motivate an extended study on how program factorization (particularly auxiliary-variable structure) impacts gradient-based learning beyond compiled NeSy methods.

\clearpage 
\section*{Acknowledgements}
We would like to thank Lorenzo Loconte, Christoph Staudt, Andreas Grivas and Joan Serrà for fruitful discussions during the writing of this paper.
Emile van Krieken was funded by the NWO AiNed project ``Human-Centric AI Agents with Common Sense'' (NGF.1607.22.044).
Luciano Serafini was funded by the
NRRP project Future AI Research (FAIR - PE00000013) under the NRRP MUR program, funded by NextGenerationEU.
Antonio Vergari was supported by the “UNREAL: Unified Reasoning Layer for Trustworthy ML” project (EP/Y023838/1) selected by the ERC and funded by UKRI EPSRC.

\paragraph{Author contributions}
The core ideas were born when SB visited the University of Edinburgh under the supervision of AV.
AV ideated approximating factors with non-deterministic circuits which SB implemented with tensor trains, further exploring product approximations. 
EvK and LS provided continuous and invaluable feedback throughout the project. 
SB conceived and implemented the experiments with guidance from AV and EvK.
EvK also helped in running baseline experiments. 
All authors contributed to writing.

\paragraph{AI use statement}
We used generative AI tools to assist in implementing the MNIST arithmetic and CLEVR experiments, for plotting, for designing the groundtruth dynamic programming solutions in weighted inference (see \cref{app:wmc_evaluation}), for drafting some of the appendices, and for reviewing the manuscript before submission. 
The main manuscript was entirely written by the authors.
\apricot was also manually implemented by the authors, only using AI for reviewing the code.
All generated content was reviewed by the authors.
We take responsibility for the final content of this work, including text and artifacts produced with the aid of generative AI.

\bibliographystyle{plainnat}
\bibliography{referomnia,reno95}

\clearpage
\appendix
\section{Circuits}
\label{app:circuits}
\subsection{Background}
We first define circuits abstractly over a semiring, which allows us to distinguish the logical and probabilistic representations used in NeSy AI.

\begin{definition}[Circuit \cite{choi2020pc,vergari2021compositional,kimmig2017algebraic}]
Let $(\mathcal{S},\oplus,\otimes,0,1)$ be a commutative semiring.
A \emph{circuit} $f(\bz)$ is a directed acyclic computational graph whose intermediate units compute $\oplus$ and $\otimes$ operations over tractable input functions of variables $\bz$.
\end{definition}

Different choices of semiring recover the circuit representations commonly used in probabilistic inference and knowledge compilation.
\emph{Probabilistic circuits} (PCs) \citep{darwiche2003differential,vergari2019tractable,choi2020pc} operate over the non-negative real semiring $(\mathbb{R}_{\geq 0},+,\times,0,1)$ and encode non-negative functions; their sum-unit weights are typically constrained to be non-negative \citep{loconte2024subtractive,loconte2024relationship}.
In contrast, \emph{logical circuits} are conventionally defined over the Boolean semiring $(\{0,1\},\lor,\land,0,1)$ as DAGs whose internal units are logical OR/AND operations and whose inputs represent logical literals \citep{darwiche2002knowledge}.

Tractable computation of multiplication and marginalization over arithmetic circuits requires certain structural properties, which we briefly discuss next.
We refer the reader to \citet{choi2020pc} for details.

We focus on circuits that are \emph{smooth}, i.e., all inputs of sum units are defined over the same variables, and \emph{decomposable}, i.e., inputs of product units are defined over disjoint sets of variables. These properties enable tractable marginalization in time linear in the circuit size \citep{choi2020pc}.

For tractable circuit multiplication, stronger structural constraints are required. A decomposable circuit is \emph{structured-decomposable} if its product units respect a common hierarchical partition of the variables, typically represented by a vtree. Two structured-decomposable circuits are \emph{compatible} \citep{vergari2021compositional} when their decompositions agree on their shared variables. Compatible circuits can be multiplied in time polynomial in their sizes while preserving smoothness and decomposability.

\subsection{Probabilistic NeSy reasoning}
To use a logical circuit together with a PC, we first cast the logical circuit into the real sum-product semiring.
This conversion is sound for \emph{deterministic} logical circuits, whose OR units have children with disjoint supports.
For such circuits, at most one child of each OR unit evaluates to $1$ for any assignment, and hence Boolean OR can be replaced by real addition and AND by multiplication without changing the encoded Boolean function.
The resulting arithmetic circuit has unit weights equal to $1$ at the former OR units.

Consequently, if $\phi$ is represented by a smooth, decomposable, deterministic logical circuit, its arithmetic representation is compatible with the PC $p_\btheta(\bc|\bx)$, and the two can be multiplied tractably, then \cref{eq:conditional_wmc} can be computed exactly and efficiently.
A first requirement is therefore constructing from $\phi$ a logical circuit with the required structural properties, a task known as \hlkc{\textbf{knowledge compilation}} (KC) and which is at least NP-hard \citep{darwiche2002knowledge,chavira2008wmc}.

Most widely used KC software operates on propositional formulas and outputs circuits over Boolean variables.\footnote{
This Boolean encoding is a property of the knowledge-compilation tools considered here, rather than an intrinsic limitation of circuits.
Circuits can in principle be defined directly over finite-domain categorical variables, for example through multi-valued decision diagrams or suitable categorical input functions.
When using propositional KC software, however, a categorical variable $c_i$ can be encoded using Boolean indicators $b_{i,j}=\mathbbm{1}[c_i=j]$ together with constraints enforcing that exactly one indicator is true.
The size increase depends strongly on the chosen encoding.
With the direct one-hot encoding used in our experiments, a $9\times9$ Sudoku changes from $d=81$ categorical variables and $N=810$ factors to $d=729$ Boolean variables and $N=10287$ factors after propositionalization and addition of mutual-exclusivity constraints.
}
Thus, the propositionalization overhead reported here reflects our use of available Boolean KC implementations, rather than a fundamental requirement of circuit representations.

Sentential decision diagrams (SDDs) \citep{darwiche2011sdd} are a popular class of deterministic and structured-decomposable logical circuits for KC.
After interpreting their deterministic OR/AND units in the real sum-product semiring, they can be combined with compatible PCs $p_\btheta(\bc|\bx)$.
They can also be constructed bottom-up by \hlexact{\textbf{multiplying}} circuit representations of factors $\phi_j$.

Let $\hlkc{\KC}(\phi_j)$ denote the arithmetic circuit obtained by compiling factor $\phi_j$ into a deterministic logical circuit and interpreting it in the sum-product semiring.
We rewrite the WMC using the symbolic operator $\hlexact{\MUL}$, which takes a set of compatible circuits as input and returns a circuit representing their product:
\begin{equation}
\sum\nolimits_{\bc \in \cC}
p_\btheta(\bc|\bx)
\hlexact{\MUL}
\big(
\hlkc{\KC}(\phi_1),\dots,\hlkc{\KC}(\phi_N)
\big)(\bc).
\end{equation}

This is the classical recipe underlying many NeSy approaches, including semantic loss \citep{xu2018semantic} and DeepProbLog \citep{manhaeve2018deepproblog}.

\section{Related Work}
\label{app:related_work}
\paragraph{Knowledge compilation} Compilers more permissive than SDDs handle larger formulas at the cost of compositionality with other circuits. d4~\cite{lagniez_2017_d4} is an exact compiler for d-DNNF, a broader class than SDDs that drops structured-decomposability.
It scales to formulas no SDD compiler can handle (e.g., 6x6 Sudoku, though still not 9x9), but the resulting circuits can only be weighted by fully-factorized probabilities, an independence assumption known to lack expressivity in practice~\cite{ahmed2022semantic,vankrieken2024independence,vankrieken2025independenceRS}. 
\citet{dierckx2025learning} propose an approximate variant of d-DNNF compilation, which enumerates a budget-limited set of models into a DNF to derive lower- and upper-bound circuits.
While these provide provable intervals, the resulting circuits are again not structured-decomposable, making them orthogonal to our target.
\citet{onaka2025ttobdd} treat TTs themselves as a target for \hlkcb{exact factor compilation}, using them in the knowledge-compilation map alongside NNF subsets and showing they are strictly more succinct than OBDDs while supporting the same polytime operations.
This corresponds to combining TT factors via exact multiplications without \hlfactorb{TT-SVD approximation} or \hlprodb{product recompression}, and inherits the same exponential blow-up under multiplication that motivates the approximate strategies we study.

\paragraph{Scalable inference in NeSy} Other approaches avoid circuit compilation entirely. 
A-NeSI~\cite{vankrieken2022anesi} trains a \emph{neural network} to amortize WMC inference, which are more expressive than circuits.
Like \apricot (but unlike SDD pipelines), A-NeSI can also learn the logic of black-box programs using input-output examples.
However, the neural approach offers no formal error guarantees and gives a representation that does not support downstream tractable operations such as multiplication with other circuits or conditioning on evidence. 
\emph{Monte Carlo methods}~\cite{maene-hardness_2024,smet2023neural,smet2023differentiable,verreet2024explain,vankrieken2025nesydiffusion} sample concept assignments to estimate gradients, which also extends to continuous settings, but \citet{maene-hardness_2024} show that the underlying gradient estimation from samples is itself \#P-hard and can require an exponential number of samples to converge.
Naturally, they also do not produce circuits that can be reused for downstream operations, targeting a different use case than ours.

\section{Error bounds}
\label{app:error_bounds}
We investigate how the different approximations impact the final WMC error.
The \hlindepb{independence} approximation is crude, with an unbounded error.
This approximation depends not only on the program semantics but also on how it is represented, which can lead to unpredictable behavior.
As we will see in our results and as we derive analytically in \cref{app:indep_categorical_binary}, the independence approximation for Sudoku derives estimates that vary by many orders of magnitude depending on the encoding used (categorical or binary).

The other approximations have more controlled error propagation.
First, the $\epsilon$ error bound on the \hlfactorb{factor} approximation introduced by \TTSVD in \apricot pipelines propagates linearly with the number of factors when \hlexactb{multiplying them exactly}.
\begin{theorem}
    \label{thm:eps_mult}
    Let $\bcX^* = \bcX_j * \bcX_l$ the element-wise product of two tensors. Let $\approxhat{\bcX^*} = \approxhat{\bcX_j} * \approxhat{\bcX_l}$, where $\|\bcX_j-\approxhat{\bcX_j}\|_F \leq \epsilon_{j}$ and $\|\bcX_l-\approxhat{\bcX_l}\|_F \leq \epsilon_{l}$. Then, $\|\bcX^*-\approxhat{\bcX^*}\|_F \le \|\bcX_j\|_\infty\,\epsilon_{l} + \|\bcX_l\|_\infty\,\epsilon_{j} + \epsilon_{j}\epsilon_{l}$
    where $\|\bcX_j\|_\infty = \max_{c_1, \dots, c_d} |\bcX_{j}(c_1,\dots,c_d)|$ is the max norm.
\end{theorem}
\begin{proof}
Let $\Delta_j = \approxhat{\bcX_j} - \bcX_j$ and $\Delta_l = \approxhat{\bcX_l} - \bcX_l$, so that $\|\Delta_j\|_F \leq \epsilon_{j}$ and $\|\Delta_l\|_F \leq \epsilon_{l}$. Expanding the product,
\begin{align*}
    \bcX^* - \approxhat{\bcX^*} 
    &= \bcX_j * \bcX_l - (\bcX_j + \Delta_j) * (\bcX_l + \Delta_l) \\
    &= -\bcX_j * \Delta_l - \Delta_j * \bcX_l - \Delta_j * \Delta_l.
\end{align*}
By the triangle inequality,
\begin{equation*}
    \|\bcX^* - \approxhat{\bcX^*}\|_F 
    \leq \|\bcX_j * \Delta_l\|_F + \|\Delta_j * \bcX_l\|_F + \|\Delta_j * \Delta_l\|_F.
\end{equation*}
For any two tensors $\bcA, \bcB$ of matching shape, the Hadamard product satisfies $\|\bcA * \bcB\|_F \leq \|\bcA\|_\infty \|\bcB\|_F$ -- follows from $\sum_\bc (\bcA(\bc) \bcB(\bc))^2 \leq \max_\bc \bcA(\bc)^2 \sum_\bc \bcB(\bc)^2 = \|\bcA\|_\infty^2 \|\bcB\|_F^2$.
Applied to the first two terms,
\begin{align*}
    \|\bcX_j * \Delta_l\|_F &\leq \|\bcX_j\|_\infty \|\Delta_l\|_F \leq \|\bcX_j\|_\infty\, \epsilon_{l}, \\
    \|\Delta_j * \bcX_l\|_F &\leq \|\bcX_l\|_\infty \|\Delta_j\|_F \leq \|\bcX_l\|_\infty\, \epsilon_{j}.
\end{align*}
For the third term, the $L^p$-norm inequality $\|\Delta_l\|_\infty \leq \|\Delta_l\|_F$ (treating the tensor as a vector) gives
\begin{equation*}
    \|\Delta_j * \Delta_l\|_F \leq \|\Delta_l\|_\infty \|\Delta_j\|_F \leq \epsilon_{l}\, \epsilon_{j}.
\end{equation*}
Summing the three bounds yields the claim.
\end{proof}

As we study boolean tensors, small $\epsilon_j$ implies $\|\bcX_j\|_\infty \approx 1$ and thus the error on $\bcX^*$ is $\cO(\epsilon_i+\epsilon_j)$.
By extension, for $N$ factors, the error on $\bcX^*$ is $\cO(N\epsilon)$ where $\epsilon = \max_j \epsilon_j$.

\hlprodb{Recompression after multiplication} (\cref{sec:tt_approx_mult}) introduces additional errors specific to each strategy.
Deterministic \TTround, in particular, can maintain guarantees on the Frobenius error of the recompressed product, which adds to the error of the exact result~\cite{oseledets2011tensor}.

Reusing \cref{thm:eps_mult} and considering the weight tensor $\bcP$ such that $\bcP(\bc)=p_\btheta(\bc\mid\bx)$, we first bound the error before the final contraction.
\begin{corollary}[Weighted-tensor error]
    \label{cor:weighted_tensor_frobenius}
    Let $\approxhat{\bcX}$ be an $\epsilon$-approximation of $\bcX$, and let $\bcP$ be a tensor of weights. Then
    \begin{equation}
        \|\bcP * \approxhat{\bcX}-\bcP * \bcX\|_F
        \leq \|\bcP\|_\infty\epsilon.
    \end{equation}
\end{corollary}

This controls the full weighted tensor and therefore applies before any marginalization or contraction. 
Since the WMC is the contraction of this tensor to a scalar, we can derive a separate bound directly. 
For tensors of matching shape, let us denote $\langle\bcA,\bcB\rangle_F=\sum_\bc\bcA(\bc)\bcB(\bc)$ for their Frobenius inner product.

\begin{corollary}[WMC error]
    \label{cor:wmc_frobenius}
    Let $\approxhat{\bcX}$ be an approximation of the constraint tensor $\bcX$ such that $\|\approxhat{\bcX}-\bcX\|_F\leq\epsilon$, and let $\bcP$ be any tensor of nonnegative weights. Then
    \begin{equation}
        \left|\wmchat-\wmc\right|
        \leq \epsilon\,\|\bcP\|_F.
        \label{eq:app_wmc_frobenius_bound}
    \end{equation}
    If $\bcP$ is a normalized probability tensor, then $\|\bcP\|_F\leq1$ and consequently $|\wmchat-\wmc|\leq\epsilon$.
\end{corollary}
\begin{proof}
Let $\Delta=\approxhat{\bcX}-\bcX$. Since the exact and approximate weighted model counts are $\langle\bcP,\bcX\rangle_F$ and $\langle\bcP,\approxhat{\bcX}\rangle_F$, respectively,
\begin{align*}
    \left|\wmchat-\wmc\right|
    &=\left|\langle\bcP,\Delta\rangle_F\right| \\
    &\leq \|\bcP\|_F\,\|\Delta\|_F
    \leq \epsilon\,\|\bcP\|_F,
\end{align*}
where the first inequality is Cauchy-Schwarz. And, if $\bcP$ is normalized, we have $\|\bcP\|_F\leq\|\bcP\|_1=1$.
\end{proof}

\section{\CTSketch and Program Dependencies}
\label{app:ctsketch}

\CTSketch~\cite{choi_ctsketch_2025} approximates each factor $\phi_j$ with a tensor train (\cref{sec:tt_factor_approx}), computes the factor WMC independently, and multiplies them progressively following a ``tree-like`` structure.
We formalize this using factor graphs and show that the tree-like assumption is violated by many NeSy programs, including the Sudoku tasks explored by \citet{choi_ctsketch_2025}, where we find that \CTSketch introduces independence approximations (\cref{eq:indep}).

\subsection{Tree-like factorizations}
\begin{example}[$\sumfour$~\cite{choi_ctsketch_2025}]
\label{ex:sumfour}
Consider the program $\sumfour$ which takes as input four digits $c_1,c_2,c_3,c_4 \in \{0,\dots,9\}$ and outputs their sum $y = c_1+c_2+c_3+c_4$.
This program can be decomposed using two intermediate sums $y_1 = c_1 + c_2$ and $y_2 = c_3 + c_4$, followed by a final addition $y = y_1 + y_2$.
The constraint factorizes as
$$
    \phi(\bc, y) = \sum_{y_1,y_2} \underbrace{\mathbbm{1}[c_1 + c_2 = y_1]}_{\phi_1(c_1,c_2,y_1)} \underbrace{\mathbbm{1}[c_3 + c_4 = y_2]}_{\phi_2(c_3,c_4,y_2)} \underbrace{\mathbbm{1}[y_1 + y_2 = y]}_{\phi_3(y_1,y_2,y)},
$$
with the corresponding factor graph shown in \cref{fig:fg-sum4}.

Let $\bc_1 = \{c_1, c_2\}$ and $\bc_2 = \{c_3, c_4\}$ denote the two disjoint subsets of $\bc$ associated with $\phi_1$ and $\phi_2$ respectively.
The WMC rewrites as:
$$
p_\btheta( y | \bx) =
\sum_{\bc} p_\btheta(\bc|\bx) \left( \sum_{y_1,y_2} \phi_1(\bc_1,y_1)\, \phi_2(\bc_2,y_2)\, \phi_3(y_1,y_2,y) \right).
$$
Assuming $p_\btheta(\bc|\bx) = \prod_i p_\btheta(c_i|\bx)$, \CTSketch simplifies this marginalization in two steps. 
\textbf{First}, since $\phi_3$ does not depend on $\bc$ and $p_\btheta(\bc|\bx)$ does not depend on $y_1, y_2$, the two sums commute and $\phi_3$ can be moved outside:
\begin{align*}
    p_\btheta( y | \bx) &=
\sum_{y_1,y_2} \phi_3(y_1, y_2, y) \left( \sum_{\bc} p_\btheta(\bc|\bx)\phi_1(\bc_1,y_1)\, \phi_2(\bc_2,y_2) \right)\\
&= \sum_{y_1,y_2} \phi_3(y_1, y_2, y) \left( \sum_{\bc} p_\btheta(\bc|\bx) \prod_{j=1}^{2} \phi_j(\bc_j,y_j) \right)
\end{align*}

\textbf{Second}, the tree structure of the factor graph and disjoint scopes of $\phi_1,\phi_2$ let us push inner marginalizations down to the leaves.
$$
p_\btheta(y|\bx) = \sum_{y_1, y_2} \phi_3(y_1, y_2, y) \prod_{j=1}^{2}
\Big(\sum_{\bc_j} p_\btheta(\bc_j|\bx)\, \phi_j(\bc_j, y_j)\Big)
$$
\end{example}

\textbf{Read in isolation, this second step has exactly the form of the \hlindepb{independence approximation} of \cref{eq:indep}}: it computes a per-factor WMC and multiplies the results.
The crucial difference is that here the equality is \emph{exact} because the leaf $\bc$-scopes are disjoint and the marginals are genuinely independent.

\CTSketch additionally approximates each $\phi_j$ as a tensor train, but the same factorization trick applies directly to the dense factor tensors $\bcX_j$.
\MNISTAdd~\cite{manhaeve2018deepproblog} satisfies this tree-like assumption, as do other programs that decompose into trees of intermediate variables.

\subsection{Overlapping factors}

\begin{figure}
\centering
\begin{subfigure}{.45\linewidth}
  \centering
  \includegraphics[width=.8\linewidth]{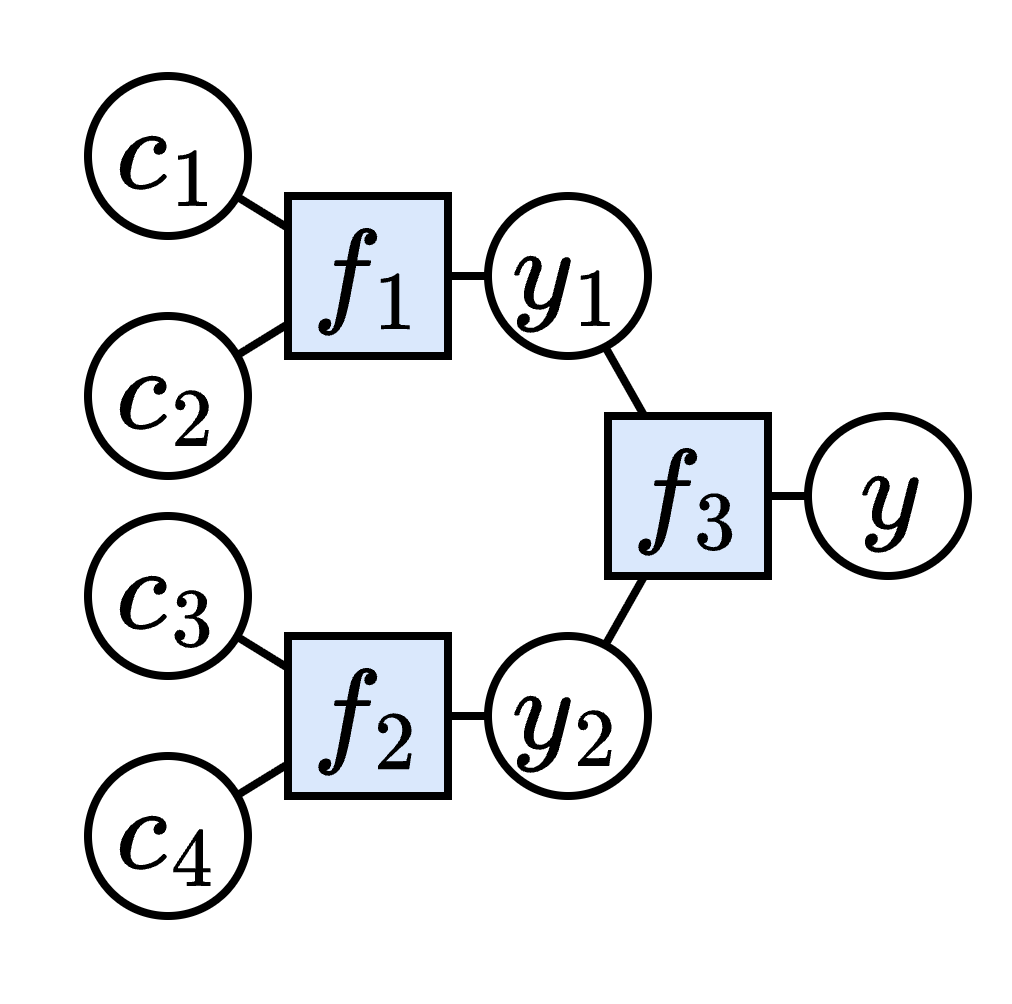}
  \caption{$\sumfour$ program.}
  \label{fig:fg-sum4}
\end{subfigure}%
\begin{subfigure}{.3\linewidth}
  \centering
  \includegraphics[width=.55\linewidth]{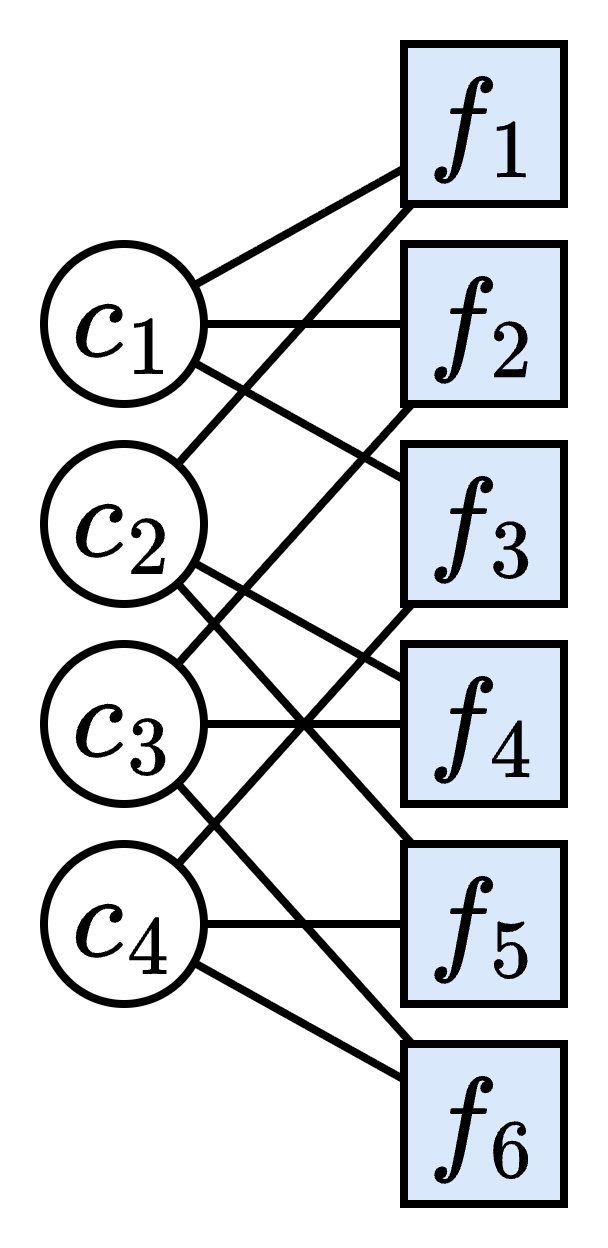}
  \caption{First-row constraint of a $4\times 4$ Sudoku program.}
  \label{fig:fg-sudoku4row}
\end{subfigure}
\caption{Factor graphs for (a) a tree-like program where marginalization sums can be pushed inside individual factors, and (b) a program with overlapping scopes and cycles where this simplification does not apply.}
\label{fig:ctsketch_fg}
\end{figure}

In most NeSy programs the factor graph is not a tree: variables appear in multiple factors and the corresponding scopes overlap substantially.
\begin{example}[Sudoku 4x4]
\label{ex:sudoku4row}
Consider the $4\times 4$ Sudoku classification task, where the program takes as input $16$ concepts $c_i \in \{1,2,3,4\}$ representing the digits in each cell and outputs whether the grid is valid.
The program decomposes into $56$ pairwise inequality constraints across rows, columns and blocks.
For example, the first row alone factorizes as
\begin{align*}
    \phi_\text{row1}(c_1,c_2,c_3,c_4) = &\underbrace{\mathbbm{1}[c_1 \neq c_2]}_{\phi_1} \underbrace{\mathbbm{1}[c_1 \neq c_3]}_{\phi_2} \underbrace{\mathbbm{1}[c_1 \neq c_4]}_{\phi_3} \\
    &\underbrace{\mathbbm{1}[c_2 \neq c_3]}_{\phi_4} \underbrace{\mathbbm{1}[c_2 \neq c_4]}_{\phi_5} \underbrace{\mathbbm{1}[c_3 \neq c_4]}_{\phi_6}.
\end{align*}
The corresponding factor graph is shown in \cref{fig:fg-sudoku4row}.
Each variable already participates in three factors within a single row, and in even more once columns and blocks are considered.
These overlapping scopes and cycles in the factor graph prevent us from pushing marginalization sums inside the product, as we did in \cref{ex:sumfour}.
\end{example}

Despite the overlap, \CTSketch applies the same per-factor simplification to Sudoku, pushing each marginalization sum to its associated factor as if the leaves had disjoint $\bc$-scopes:
\begin{equation*}
    \sum_\bc p_\btheta(\bc|\bx) \prod_{j} \phi_j(\bc_j) \stackrel{\indep,\fac}{\approx} \hlindep{\prod_j} \Big(\sum_{\bc_j} p_\btheta(\bc_j|\bx)\, \hlfactor{\operatorname{AC}}(\phi_j)(\bc_j)\Big).
\end{equation*}
The right-hand side has exactly the form of the (exact) factorization in \cref{ex:sumfour}, but here the overlapping $\bc_j$ turn the equality into an approximation: it is the composition of a \hlfactorb{factor approximation} -- each $\phi_j$ replaced by its TT $\hlfactor{\operatorname{AC}}(\phi_j)$ -- with an \hlindepb{independence approximation}.
The independence step is the source of the unbounded error analyzed in \cref{app:error_bounds} and empirically illustrated in \cref{app:indep_categorical_binary}.

\section{Independence Approximation and Categorical vs. Binary Encoding}
\label{app:indep_categorical_binary}

We illustrate how the \hlindepb{independence approximation} can yield drastically different model count estimates depending on whether $\phi$ is encoded over categorical or binary variables, even though the two encodings describe the same set of valid worlds. We use Sudoku as a concrete example, with the same constraint expressed once over categorical cells and once after the propositionalization required by SDD compilation (\cref{sec:exact_circuit_compilation}).

\subsection{Independence Approximation under a Uniform Prior}
Specializing the independence approximation to a uniform prior $p_\btheta(\bc) = 1/|\cC|$, the WMC reduces to the model count $\text{MC} = \sum_\bc \phi(\bc)$, and the approximation becomes
\begin{equation}
    \label{eq:indep_mc_uniform}
    \widehat{\text{MC}}^{\,\indep} 
    \;=\; |\cC| \cdot \hlindep{\prod_{j=1}^N} p_j, 
    \qquad p_j \;:=\; \frac{1}{|\cC_j|}\sum_{\bc_j \in \cC_j} \phi_j(\bc_j),
\end{equation}
where $p_j$ is the fraction of assignments over the scope $\bc_j$ that satisfy $\phi_j$. Both factors of \cref{eq:indep_mc_uniform} depend on the encoding only through trivial counting, but their product is dominated by very different magnitudes in the two encodings.

\subsection{Sudoku 4x4: Categorical Encoding}
The categorical formulation has variables $c_i \in \{1,\dots,4\}$ for $i=1,\dots,16$, giving $|\cC| = 4^{16}$. The constraint $\phi$ decomposes into $N=56$ pairwise inequalities $\phi_j(c_i,c_l) = \mathbbm{1}[c_i \neq c_l]$ over cell pairs $(i,l)$ sharing a row, column, or box. Each factor has $|\cC_j|=16$ joint assignments of which $12$ are satisfying, so $p_j = 3/4$. Plugging into \cref{eq:indep_mc_uniform},
\begin{equation*}
    \widehat{\text{MC}}^{\,\indep}_{\text{cat}} 
    \;=\; 4^{16} \cdot \left(\tfrac{3}{4}\right)^{56} 
    \;\approx\; 432.9,
\end{equation*}
which overestimates the true MC of $288$ by a factor of $\approx 1.5\times$.

\subsection{Sudoku 4x4: Binary Encoding}
SDD compilation requires propositionalizing $\phi$ via indicators $b_{i,v} = \mathbbm{1}[c_i = v]$ for $i=1,\dots,16$ and $v \in \{1,\dots,4\}$, resulting in $64$ Boolean variables and $|\cC| = 2^{64}$. The constraint becomes a conjunction of $N = 336$ CNF clauses of four types:
\begin{itemize}
    \item \emph{At-least-one} ($16$ clauses): $b_{i,1} \lor b_{i,2} \lor b_{i,3} \lor b_{i,4}$ per cell, with $p_j = 1 - (1/2)^4 = 15/16$.
    \item \emph{At-most-one} ($96$ clauses): $\lnot b_{i,v} \lor \lnot b_{i,v'}$ for each cell $i$ and value pair $v < v'$, with $p_j = 3/4$.
    \item \emph{Distinct values across rows and columns} ($96 + 96$ clauses): $\lnot b_{i,v} \lor \lnot b_{l,v}$ for each row- or column-pair $(i,l)$ and each value $v$, with $p_j = 3/4$.
    \item \emph{Distinct values across boxes} ($32$ clauses): same form, restricted to box-pairs not already covered by rows or columns, with $p_j = 3/4$.
\end{itemize}
Plugging into \cref{eq:indep_mc_uniform},
\begin{equation*}
    \widehat{\text{MC}}^{\,\indep}_{\text{bin}} 
    \;=\; 2^{64} \cdot \left(\tfrac{15}{16}\right)^{16} \cdot \left(\tfrac{3}{4}\right)^{96+96+96+32}
    \;\approx\; 6.87 \times 10^{-22},
\end{equation*}
underestimating the true MC of $288$ by roughly $24$ orders of magnitude.

\subsection{Discussion}

The two estimates differ by approximately $24$ orders of magnitude despite describing the same $288$ valid boards. It is tempting to attribute this entirely to the $N$ scaling of \cref{eq:indep_mc_uniform}, as propositionalization inflates the factor count from $56$ to $336$, and each additional sub-unit $p_j$ pushes the product toward zero. Our inference experiments in \cref{sec:experiments} show that this picture is misleading: even when the SDD pipeline reduces the propositionalized factors back to a count $\hlindep{K}$ comparable to the categorical $N$ via exact multiplications, the binary estimate continues to underestimate the true MC while the categorical one continues to overestimate it, and the two remain orders of magnitude apart.

The deeper issue is that the independence approximation, viewed as a fuzzy semantics with the product t-norm, is unbounded and \emph{syntax-dependent}: its value depends not only on the set of models of $\phi$ but on how $\phi$ is written. As a minimal illustration, consider a single Boolean $a \in \{0,1\}$. The formulas $\phi_1 := a \land a$ and $\phi_2 := a$ have the same single satisfying assignment ($\text{MC}=1$), yet $\widehat{\text{MC}}^{\,\indep}(\phi_2) \;=\; 2 \cdot \tfrac{1}{2} \;=\; 1$ and $\widehat{\text{MC}}^{\,\indep}(\phi_1) \;=\; 2 \cdot \tfrac{1}{2} \cdot \tfrac{1}{2} \;=\; \tfrac{1}{2}$.

Categorical and binary encodings of Sudoku express the same underlying constraint with very different syntactic profiles, and the product t-norm amplifies these differences in opposite directions: even after multiplying propositionalized factors back together, the residual factor structure is not the same as the categorical one and produces a systematically smaller $\prod_j p_j$.
This contrasts with fuzzy semantics built on other t-norms (Gödel, Łukasiewicz), which have more established bounds and structural guarantees.~\cite{vankrieken2022fuzzy} Neither, however, corresponds to a probabilistic relaxation of \cref{eq:wmc}, which is what motivates the use of the product t-norm as the \hlindepb{independence approximation} in the first place.
\section{Tensor Train Algebra}
\label{app:tt_algebra}
\begin{figure}
    \centering
    \begin{subfigure}[t]{0.32\linewidth}
        \centering
        \includegraphics[width=\linewidth]{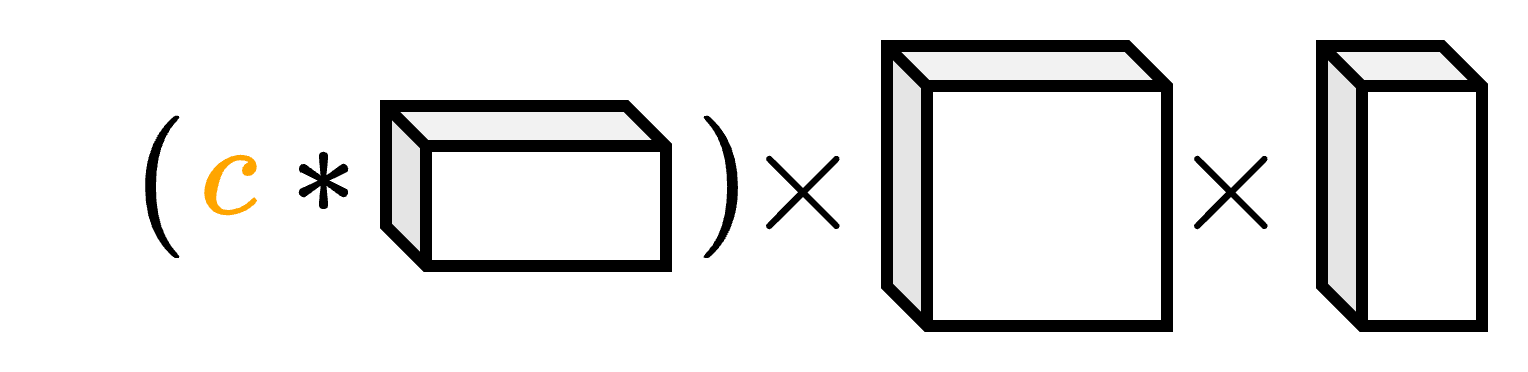}
        \caption{$\hlexact{c *} \bcX$}
        \label{fig:tt-algebra-app-scalar}
    \end{subfigure}
    \hfill
    \begin{subfigure}[t]{0.32\linewidth}
        \centering
        \includegraphics[width=\linewidth]{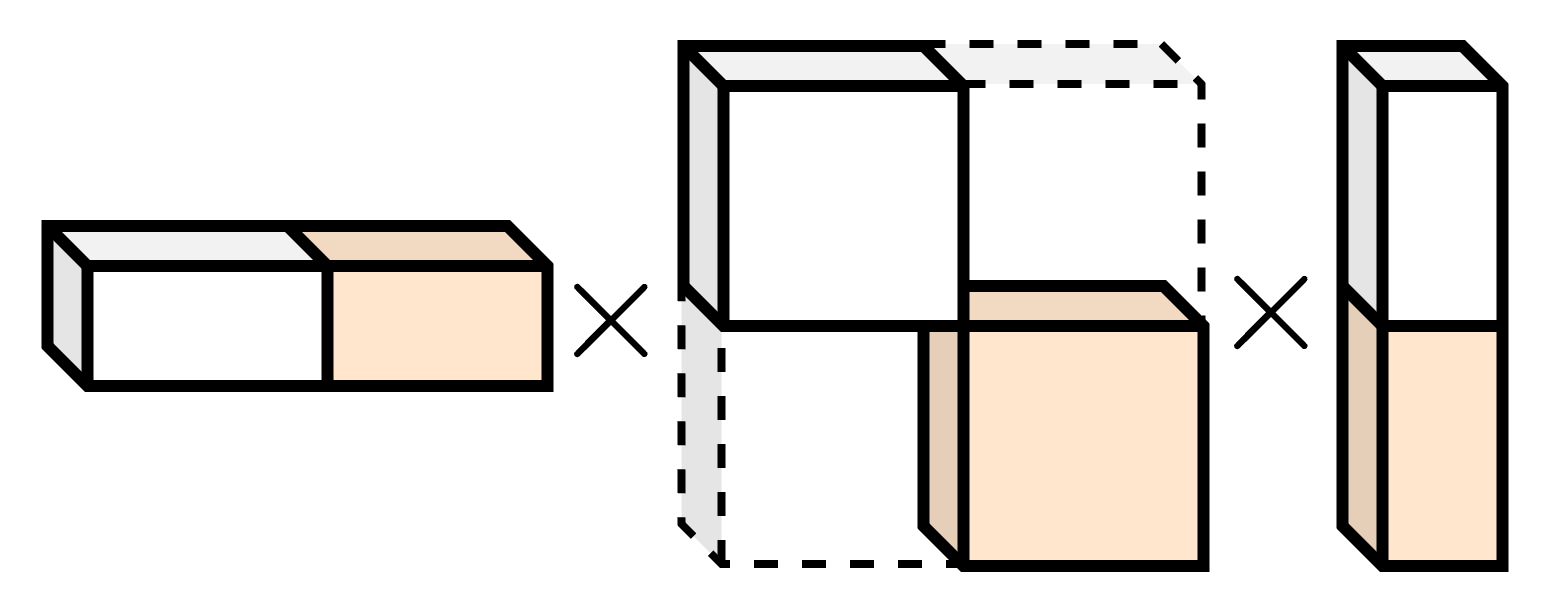}
        \caption{$\bcX_j \hlexact{ + \bcX_l}$}
        \label{fig:tt-algebra-app-add}
    \end{subfigure}
    \hfill
    \begin{subfigure}[t]{0.32\linewidth}
        \centering
        \includegraphics[width=.7\linewidth]{assets/tt-algebra-3.png}
        \caption{$\bcX_j \hlexact{ * \bcX_l}$}
        \label{fig:tt-algebra-app-mul}
    \end{subfigure}

    \vspace{1em}

    \begin{subfigure}[t]{0.48\linewidth}
        \centering
        \includegraphics[width=.7\linewidth]{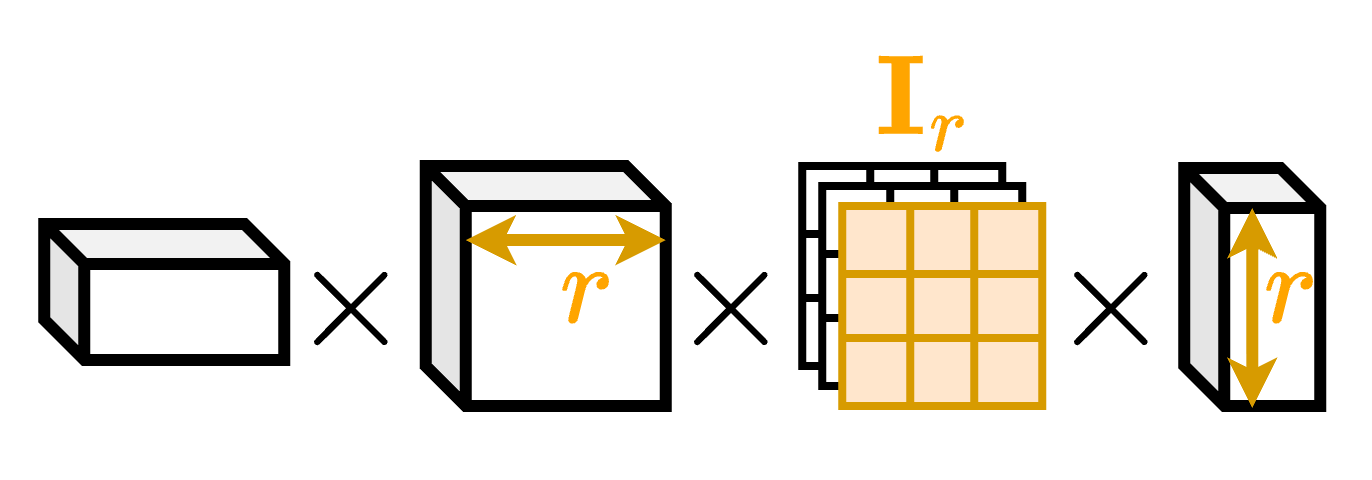}
        \caption{$\operatorname{Broadcast}_{ c_1 c_2 c_3 \to c_1 c_2 \hlexact{c_4} c_3}$ with cost $\cO(nr^2)$ }
        \label{fig:tt-algebra-app-broadcast}
    \end{subfigure}
    \hfill
    \begin{subfigure}[t]{0.48\linewidth}
        \centering
        \includegraphics[width=.7\linewidth]{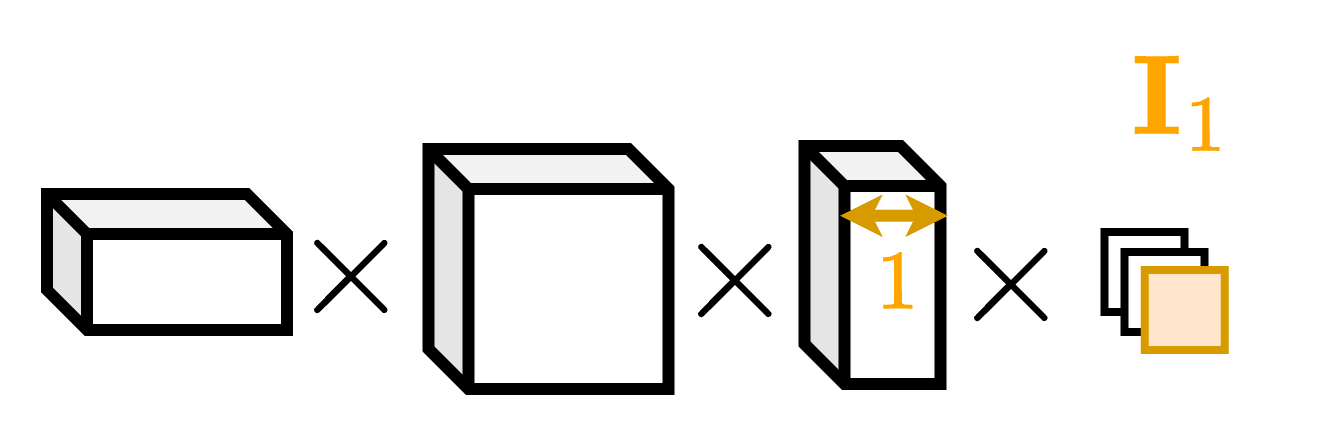}
        \caption{$\operatorname{Broadcast}_{ c_1 c_2 c_3 \to c_1 c_2  c_3 \hlexact{c_4}}$ with cost $\cO(n)$ }
        \label{fig:tt-algebra-app-broadcast-efficient}
    \end{subfigure}
    \caption{Closed algebra over tensor trains: (a) scalar multiplication, (b) addition, (c) Hadamard product, and (d-e) broadcasting with two different insertion positions. The cost of broadcasting is given as the size of the inserted \emph{identity cores}.}
    \label{fig:tt-algebra-app}
\end{figure}
\subsection{Addition, Scalar Multiplication and Hadamard Product}

We briefly introduce the closed-form algebra of TTs~\cite{oseledets2011tensor}. \cref{fig:tt-algebra-app} provides a graphical illustration. Let $\bcA$ and $\bcB$ denote two TTs over the same $d$ variables in the same core-to-variable ordering, with cores $\bcT_{\bcA,k}$ and $\bcT_{\bcB,k}$ of TT ranks $r^{A}_k$ and $r^{B}_k$ respectively. We write $n = \max_k n_k$ for the maximal mode size and assume for simplicity that input TTs share a common rank $r$.

\paragraph{Scalar multiplication (\cref{fig:tt-algebra-app-scalar})} For $c \in \R$, the TT $c \cdot \bcA$ is obtained by absorbing $c$ into any single core, e.g. $\bcT_{c\bcA, 1}(c_1) = c\, \bcT_{\bcA, 1}(c_1)$. The ranks are unchanged and the resulting TT has the same $\cO(d n r^2)$ storage as $\bcA$.

\paragraph{Addition (\cref{fig:tt-algebra-app-add})} The sum $\bcA + \bcB$ is a TT whose cores stack the input cores block-diagonally,
\[
    \bcT_{\bcA+\bcB,k}(c_k) = \begin{pmatrix} \bcT_{\bcA,k}(c_k) & 0 \\ 0 & \bcT_{\bcB,k}(c_k) \end{pmatrix}, \qquad 1 < k < d,
\]
with the boundary cores stacked horizontally ($k=1$) and vertically ($k=d$) so that $r_0 = r_d = 1$ is preserved. The TT ranks add: $r^{A+B}_k = r^A_k + r^B_k$, giving total storage $\cO(d n (r^A_k + r^B_k)^2) = \cO(d n r^2)$ for two equal-rank inputs.

\paragraph{Hadamard product (\cref{fig:tt-algebra-app-mul})} The element-wise product $\bcA * \bcB$ has cores given by the per-site Kronecker product of \cref{eq:tt-mult},
\[
    \bcT_{\bcA*\bcB,k}(c_k) = \bcT_{\bcA,k}(c_k) \otimes \bcT_{\bcB,k}(c_k),
\]
with multiplied ranks $r^{A*B}_k = r^A_k \cdot r^B_k$ and storage $\cO(d n (r^A_k r^B_k)^2) = \cO(d n r^4)$ for two equal-rank inputs. This quadratic per-product blow-up is the exponential rank growth discussed in \cref{sec:tt_exact_mult} and motivates the recompression strategies of \cref{sec:tt_approx_mult}.

\subsection{Broadcasting}
\label{app:tt_broadcasting}

The cores in \cref{eq:tt-mult} assume that $\bcX_j$ and $\bcX_l$ are TTs over (i) the same scope and (ii) the same core-to-variable ordering. Assumption (ii) is easily enforced by fixing a global ordering $\sigma$ of the variables in $\bc$ and decomposing every factor in that ordering. Assumption (i) requires \emph{broadcasting}: any TT whose scope $\bc_j \subsetneq \bc$ must be lifted to the common scope $\bc$ before it can be multiplied with another TT.

\paragraph{Identity cores}
For a TT $\bcX_j$ with scope $\bc_j$ decomposed in the ordering induced by $\sigma$, broadcasting to $\bc$ inserts, for every missing variable $c_k \notin \bc_j$, an \emph{identity core} $\bcI_k$ at the position dictated by $\sigma$:
\[
    \bcI_k(c_k) = \bI_{r_{k-1}}, \qquad c_k = 1, \dots, |\cC_k|,
\]
where $r_{k-1}$ is the rank of the bond at which $\bcI_k$ is inserted, so that $\bcI_k$ has shape $r_{k-1} \times |\cC_k| \times r_{k-1}$. Since the slices of $\bcI_k$ are constant in $c_k$, the broadcasted TT evaluates to the same value as $\bcX_j$ for every assignment of the missing variables, as required.

\paragraph{Position matters}
The position at which an identity core is inserted directly controls its size, and hence the storage and downstream multiplication cost of the broadcasted TT. If $c_k$ is inserted \emph{within the span} of $\bc_j$ in $\sigma$ -- i.e., between two original cores -- then $\bcI_k$ has size $\cO(n r^2)$, where $r$ is the local rank of $\bcX_j$ at that bond. If instead $c_k$ is inserted \emph{outside the span} -- before the first or after the last variable of $\bc_j$ -- the surrounding bond rank is $r_0 = r_d = 1$ and $\bcI_k$ collapses to a $1 \times |\cC_k| \times 1$ core of size $\cO(n)$, leaving the storage essentially unchanged. \cref{fig:tt-algebra-app-broadcast,fig:tt-algebra-app-broadcast-efficient} contrast the two cases: inserting the same variable $c_4$ between $c_2$ and $c_3$ costs $\cO(nr^2)$, while appending it after $c_3$ costs only $\cO(n)$.

\paragraph{Choosing a global variable ordering}
\def\pos{\operatorname{pos}}
\def\gaplen{\operatorname{gaplen}}
Different global orderings $\sigma$ result in different broadcasting costs because they change which missing variables fall inside or outside each factor's span. We seek a $\sigma$ that minimizes the total cost of broadcasting all factors to their pairwise (or full) common scopes.
We assume for simplicity a uniform rank $r$ and uniform mode size $n$ across all TTs.~\footnote{In practice, \TTSVD yields non-uniform ranks, but estimating the exact post-broadcast cost would require recompiling every factor for every candidate $\sigma$, which is prohibitive. The uniform-rank approximation is enough to capture the essential combinatorial structure of the problem.} 
The broadcasting cost of factor $\phi_j$ under $\sigma$ scales with its number of \emph{interior gaps}: the number of positions inside the span of $\bc_j$ that are occupied by variables outside $\bc_j$. Concretely, denoting by $\pos_\sigma(c_i)$ the position of $c_i$ in $\sigma$, the interior gap length of $\phi_j$ is
\begin{equation}
    \label{eq:gaplen}
    \gaplen_j(\sigma) = \Big( \max_{c_i \in \bc_j} \pos_\sigma(c_i) - \min_{c_i \in \bc_j} \pos_\sigma(c_i) + 1 \Big) - |\bc_j|,
\end{equation}
where the parenthesised term is the span of $\bc_j$ in $\sigma$ and $|\bc_j|$ is the number of own variables it contains. 
We seek the ordering minimising the total gap length,
\begin{equation}
    \label{eq:order_obj}
    \min_\sigma \; \sum_{j=1}^N \gaplen_j(\sigma).
\end{equation}

\paragraph{Spectral ordering}
\Cref{eq:order_obj} is a variant of the (NP-hard) minimum linear arrangement problem on the hypergraph whose hyperedges are the factor scopes $\bc_j$. We solve it heuristically via \emph{spectral seriation}, which is fast, parameter-free, and reliably produces orderings that are good enough for our pipelines.

We construct a weighted co-occurrence graph on the variables $\bc$ with adjacency
\[
    \bW_{ii'} = \sum_{j : c_i, c_{i'} \in \bc_j} 1,
\]
i.e., the number of factors in which $c_i$ and $c_{i'}$ jointly appear. Variables that co-occur in many factors should sit close together in $\sigma$ to keep factor spans short. Letting $D = \operatorname{diag}(\bW \mathbf{1})$ be the degree matrix and $L = D - \bW$ the graph Laplacian, the \emph{Fiedler vector} -- the eigenvector associated with the second-smallest eigenvalue of $L$ -- provides a one-dimensional embedding of the variables that approximately minimises the weighted sum of squared distances between adjacent vertices~\cite{atkins2003_spectral}. We sort the variables by their Fiedler-vector entries to obtain $\sigma$.

This ordering is computed once per WMC instance, before any TT is built, and is then reused for all factors.

\section{Multiplication Schedules}
\label{app:partitioning}

The bottom-up compilation pipeline of \cref{sec:pipeline} pairs factors at two distinct stages: 
(i) the optional \emph{pre-multiplication} of Step I, which groups dense factor tensors $\bcX_j$ and multiplies them in dense space before applying TT-SVD (TT pipelines only), and 
(ii) the \emph{circuit multiplication} of Step II, which iteratively pairs and multiplies the $N$ compiled factor circuits into the $K \leq N$ circuits of Step III. 
In both stages, we intuitively aim to prioritize the multiplication of factors which share the most dependencies (that would otherwise be approximated by Step III) under some space constraints. 
The factors grow differently in the different stages, and we therefore use different strategies.

\subsection{Dense Pre-multiplication: Hypergraph Partitioning}
\label{app:partitioning_premul}

Before converting factors to TTs, we may multiply several sufficiently small factors exactly as dense tensors. This reduces the number of circuits passed to Step II and ensures that dependencies among factors in the same group are represented before any TT approximation is introduced. The grouping must balance two objectives: factors sharing variables should preferably be placed together, but the dense tensor produced by each group must fit within a prescribed memory budget $B$.

\paragraph{Grouping objective}
Let $\cP_1,\ldots,\cP_K$ be a partition of the $N$ initial factors. The factors in group $\cP_k$ are replaced by their dense Hadamard product
\[
    \bcX_{\cP_k}
    =
    \prod_{j\in\cP_k}\bcX_j,
    \qquad
    \bc_{\cP_k}
    =
    \bigcup_{j\in\cP_k}\bc_j.
\]
The resulting tensor contains one mode for every variable in the union of the grouped scopes and therefore has
$\operatorname{size}(\bcX_{\cP_k})
    =
    \prod_{c_i\in\bc_{\cP_k}}|\cC_i|$
entries. We require this size to remain below $B$.

We seek groups that share as few variables as possible with one another while satisfying this constraint. 
Equivalently, factors containing the same variable should be kept in the same group whenever the memory budget permits. 

\paragraph{Hypergraph representation}
This grouping problem has a natural hypergraph representation. We create one node for each factor and one hyperedge for each variable. The hyperedge for variable $c_i$ connects exactly the factors whose scopes contain $c_i$:
\[
    e_i=\{j:c_i\in\bc_j\}.
\]
If the connected factors are assigned to different groups, then $e_i$ is cut. A cut hyperedge therefore corresponds directly to a variable shared by multiple output groups. Minimizing the number of cut hyperedges implements the dependency objective above.

\paragraph{Memory constraint}
The exact dense size of a group depends on the union of its factor scopes and is not additive across factors. Standard hypergraph partitioners, however, impose balance constraints using additive node weights. We therefore give factor $j$ the conservative weight
\[
    w_j
    =
    \log\!\left(\prod_{c_i\in\bc_j}|\cC_i|\right).
\]
For a group $\cP_k$,
\[
    \sum_{j\in\cP_k}w_j
    =
    \log\!\left(
        \prod_{j\in\cP_k}
        \prod_{c_i\in\bc_j}|\cC_i|
    \right)
\]
upper-bounds the logarithm of its actual dense size. The bound may count a shared variable more than once, whereas the dense product contains that variable only once. Consequently, enforcing
\[
    \sum_{j\in\cP_k}w_j\leq\log B
\]
is sufficient, though potentially conservative, for keeping the dense product within budget.

The resulting optimization is
\begin{equation}
    \label{eq:premul_partitioning}
    \min_{\cP_1,\ldots,\cP_K}
    \big|\{\,i:e_i\text{ is cut}\,\}\big|
    \qquad
    \text{s.t.}
    \qquad
    \sum_{j\in\cP_k}w_j\leq\log B
    \quad\forall k.
\end{equation}
Thus, the objective keeps factors coupled by shared variables together, while the constraint limits the cost of materializing each resulting dense tensor.

\paragraph{Implementation}
We approximately solve this constrained partitioning problem with KaHyPar~\citep{schlag2021kahypar}. Starting from $K=1$, we increase the number of groups until the partition satisfies the dense-size budget. We use $B\in\{10^6,10^9\}$ entries in our experiments. The partitioning overhead is negligible relative to dense multiplication and circuit compilation.

\subsection{Circuit Multiplication Schedules}
\label{app:partitioning_mul}

After Step I, the pipeline contains $N$ compiled circuits, which Step II reduces to $K$ through pairwise multiplications. The multiplication schedule balances two objectives. First, factors with important shared dependencies should be combined before the pipeline stops and approximates the remaining groups as independent. Second, intermediate products must remain small enough to store and, for recompressed TT multiplication, simple enough to approximate accurately under the available rank budget.

This scheduling problem is closely related to the \emph{contraction-order} problem studied for tensor networks~\citep{gray_tncontractorder_2021}. Standard tensor-network contraction normally continues until the entire network has been contracted, so the order is chosen primarily to minimize the sizes of intermediate tensors. 
Our setting introduces an additional concern: Step II may stop at $K>1$, after which dependencies between the remaining groups are ignored by the fuzzy reduction. The schedule must therefore decide not only how to control intermediate size, but also which dependencies should already have been incorporated when multiplication stops. 
TT recompression adds a further consideration: even when two schedules have comparable memory costs, they may expose different intermediate functions to rounding and consequently result in different approximation errors.

We explore three different scheduling strategies, each which their own focus within these objectives and considerations.
These strategies are heuristic.
Just as tensor network literature shows no single merge order strategies across problems~\cite{gray_tncontractorder_2021,gray_tnapproxcontractorder_2024}, we empirically find different strategies to perform better in different experiments.

\paragraph{Dependency-first matching}
When no more specific program structure is available, we prioritize the first objective directly. At each level of a binary multiplication tree, we pair circuits with the largest scope intersections. Scope overlap serves as a proxy for shared dependencies: multiplying highly overlapping factors first keeps those dependencies inside the same circuit if the pipeline stops early, rather than leaving them to be discarded by the final independence approximation. We choose the pairs jointly through maximum-weight matching~\cite{Edmonds2010matching}, with edge weights $w_{jl}=|\bc_j\cap\bc_l|$.
We use this strategy in our Sudoku experiments.

\paragraph{Variable-elimination scheduling}
When a program contains auxiliary variables, we can minimize its intermediate size by quickly eliminating these variables. 
Auxiliary variables encode dependencies between factors, but they are not queried and ultimately receive uniform weights. 
Once all factors containing such a variable have been multiplied, it can therefore be marginalized exactly. 
Doing so immediately can sharply and predictably reduce the scope and ranks of the intermediate circuit before the following multiplications.
The variable-elimination schedule consequently prioritizes pairs whose multiplication makes the largest number of auxiliary variables available for marginalization. 
Note that this still serves the same dependency objective: all factors coupled through an eliminated variable are combined before that variable is summed out, so its induced dependency is retained exactly within the resulting factor. 
We use this strategy in our experiments with auxiliary variables (MNIST arithmetic and CLEVR-CountDiff).

\paragraph{Ordered contraction}
Finally, we explore a strategy inspired by sequential boundary contraction in tensor-network algorithms~\citep{orus2013practical,schollwoeck2010density} which helps in intermediate TT rounding steps. 
Boundary-MPS methods progressively contract a tensor network from one side and compress the resulting boundary after each step. 
Their guiding principle is that each truncation approximates the complete network prefix contracted so far, rather than an arbitrary subset of its tensors.
We transfer this principle to factor multiplication under the variable order of the TT. 
Each factor is assigned to the last variable in its scope. 
Factors assigned to the same variable are multiplied together, and the resulting groups are incorporated from left to right into a running product. 
After processing variable $c_i$, the accumulator therefore contains every factor whose scope is contained in the prefix $(c_1,\ldots,c_i)$.
Consequently, each TT rounding acts on a complete factor prefix. Dependencies whose factors close within that prefix can interact before ranks are truncated, whereas an arbitrary binary tree may round disconnected or incomplete collections of factors and retain distinctions that later factors would make redundant. 
This schedule does not guarantee smaller ranks for every program, but it aligns intermediate approximation with the sequential decomposition underlying the TT. CLEVR-Diff uses this schedule.

\subsection{Final contraction}
Compilation generally leaves two TT factors (when feasible), which are contracted jointly with the query weights without materializing their product TT. 
This final contraction is exact and introduces no independence approximation or additional truncation. Because it avoids storing the inflated product cores, the inner product of $m$ TTs with rank $r$ is $O(r^m)$ rather than $O(r^2m)$; with two TTs stored in $O(r^2)$ memory, this final contraction of two TTs is therefore virtually free in space complexity.

\section{Experiments (cont.)}
\subsection{Conditional WMC}
\label{app:conditional_wmc}
In the conditional WMC setting, we still use a parameterized model $p_\btheta(\bc|\bx)$ over the symbolic concepts, but the logical constraint $\phi(\bc,\by): \cC \times \cY \to \{0,1\}$ now also depends on output variables $\by$. 
The goal is to use the formula to derive a probability of $\by$ given $\bx$:
\begin{equation}
    \label{eq:conditional_wmc}
    p_\btheta(\by | \bx) := \frac{\sum_{\bc \in \cC} p_\btheta(\bc|\bx) \phi(\bc, \by)}{\sum_{\by' \in \cY, \bc \in \cC} p_\btheta(\bc|\bx) \phi(\bc, \by')}
\end{equation}
which, given a label $\bl \in \cY$, is optimized by maximizing $p_\btheta(\bl | \bx)$.

The denominator is a normalizing constant that ensures $p_\btheta(\by|\bx)$ is a valid distribution over $\by$.
If, for every $\bc \in \cC$, there exists one and only one $\by \in \cY$ such that $\phi(\bc, \by) = 1$, then the denominator simplifies to $1$. 
Many works~\cite{choi_ctsketch_2025,vankrieken2022anesi,vankrieken2025nesydiffusion} take this assumption and write the symbolic program as a function $\by = f(\bc)$.
We note however that this is generally not the case for Sudoku completion: some partial boards $\bc$ could be completed in multiple ways, and some could not admit any completion.
However, the SatNet dataset~\cite{wang_2019_satnet} is constructed so that every $\bc$ has only one valid completion, such that in practice works that optimize the numerator alone~\cite{choi_ctsketch_2025} still achieve good results. 

\subsection{Sudoku compilation ablation}
\label{app:sudoku_ablation}

\subsubsection{Evaluated pipelines}
The Sudoku ablation plot in \cref{fig:pareto-inference} records the \apricot compilation pipelines using 
$$
 N\ \hlexact{\overset{p\times}{\longrightarrow}}\
 \hlfactor{\overset{R_{\mathrm{init}}}{\longrightarrow}}\
 \hlexact{\overset{e\times}{\longrightarrow}}\
 \hlprod{\overset{t\times}{\longrightarrow}}\ \hlindep{K}.
$$
Here $N$ is the initial factor count, $p$ counts dense pre-multiplications, $e$ counts exact TT merges, and $t$ counts recompressed merges. The TT-SVD arrow records the initial rank setting.
$\hlindep{K}$ is the resulting number of factors; if $\hlindep{K} > 1$, the baseline uses an independent reduction.
Note that some of these operations are optional (e.g., pre-multiplication, recompressions, ...); when they are absent, the arrow is omitted.
We test different dense pre-multiplication budgets (none, $10^6$, or $10^9$ entries, see also \cref{app:partitioning_premul}), initial ranks $\hlfactor{R_\text{init}} \in \{10, 30, 500\}$, recompression strategies from \cref{app:tt_mult_algos} with a maximum rank of $500$, and various reduction targets $\hlindep{K}$.

SDD pipelines are recorded as
$$N \
        \hlkc{\overset{\operatorname{KC}}{\longrightarrow}} \
        \hlexact{\overset{e \times}{\longrightarrow}}\ \hlindep{K}$$
where the KC arrow denotes exact compilation of the initial clauses, followed by any exact merge arrows. 
As for the TT pipelines, if $\hlindep{K} > 1$, fuzzy approximations are used.

\subsubsection{Multiplication schedule and rank budget}
\label{app:sudoku_schedule_rank_ablation}
We further ablate the multiplication schedule and maximum rank on Sudoku $6\times6$, using \TTround for recompressed TT multiplication and stopping at five factors before fuzzy reduction. The dependency-first schedule from \cref{app:partitioning_mul} prioritizes factors with overlapping scopes. Its intermediate products can therefore be more difficult to represent within the available rank budget, resulting in fewer exact merges than under a random schedule (169 rather than 183 at rank 500). Similarly, reducing the maximum rank to 50 also leaves fewer merges within the exact rank budget (144). This behavior is expected from the compilation algorithm; the relevant comparison is the accuracy and cost of the resulting pipelines.

\begin{table}[h]
    \centering
    \small
    \begin{tabular}{l l r r r}
        \toprule
        Pipeline & Schedule & Rank & Time (s) & $|\log_{10}(\mathrm{MC}/\mathrm{MC}_{\mathrm{ref}})|\downarrow$ \\
        \midrule
        $216\ \hlfactor{\overset{R=6}{\longrightarrow}}\ \hlexact{\overset{169\times}{\longrightarrow}}\ \hlprod{\overset{42\times}{\longrightarrow}}\ \hlindep{5}$
            & Dependency-first & 500 & 42.4 & 1.30 \\
        $216\ \hlfactor{\overset{R=6}{\longrightarrow}}\ \hlexact{\overset{183\times}{\longrightarrow}}\ \hlprod{\overset{28\times}{\longrightarrow}}\ \hlindep{5}$
            & Random & 500 & 94.7 & 3.41 \\
        $216\ \hlfactor{\overset{R=6}{\longrightarrow}}\ \hlexact{\overset{144\times}{\longrightarrow}}\ \hlprod{\overset{67\times}{\longrightarrow}}\ \hlindep{5}$
            & Dependency-first & 50 & 5.1 & 1.97 \\
        \bottomrule
    \end{tabular}
    \caption{Ablation of the multiplication schedule and maximum TT rank on Sudoku $6\times6$. All pipelines use \TTround and reduce the 216 initial factors to five before fuzzy reduction.}
    \label{tab:sudoku_schedule_rank_ablation}
\end{table}

At rank 500, dependency-first matching is substantially more accurate than the random schedule. Although the random schedule admits more exact merges, its five remaining factors preserve less of the shared dependency structure before fuzzy reduction. 
The dependency-first matching is also surprisingly faster, perhaps because handling dependencies early on while the TTs are still small is more efficient than in later stage.
Comparing the two dependency-first pipelines, rank 50 is much faster, consistent with the cubic dependence of TT multiplication on the rank, whereas rank 500 is more expressive and yields a more accurate model count. Notably, dependency-first matching at rank 50 remains more accurate than the random schedule at rank 500.

\subsubsection{Sudoku 4x4}
\label{app:pareto_n4}
On Sudoku 4x4, we confirm that both SDD and \apricot pipelines can derive the exact model count within a tiny $\epsilon$ error.
As SDD can exactly reduce the formula into a single circuit, we also expect that approximate compilation can achieve a reduction to $K=1$ with negligible error, which is what we observe in practice.
We also notice that, as introducing approximate multiplications is slower than exact ones due to the overhead of recompression (though, they still compile within seconds), they are not beneficial in such small scale settings, and the best point on the Pareto front is achieved by the SDD pipelines.

\begin{figure}
    \centering
    \includegraphics[width=.5\linewidth]{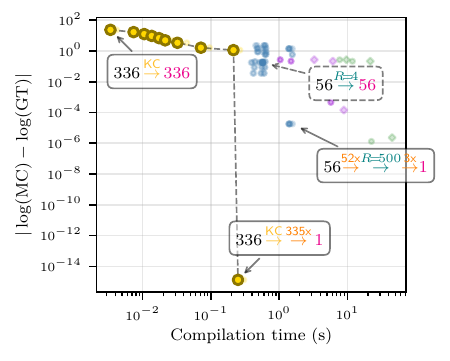}
    \caption{Pareto front of log inference error vs compilation time for Sudoku 4x4.}
    \label{fig:pareto_n4}
\end{figure}

\subsection{Neurosymbolic tasks}
\label{app:programs}
\label{app:task_details}
We describe the Visual Sudoku, arithmetic and scene-understanding tasks used in our experiments. 
Perception predicts distributions over input MNIST digits or object attributes, and supervision is provided only on the program output. Auxiliary symbolic variables express intermediate computations and are unweighted: compiled pipelines marginalize them jointly, while independence reduction marginalizes them separately within each factor.

\subsubsection{Definitions}
\label{app:constraints}
\paragraph{ViSudo Classification} 
Let $z_{r,c} \in \{1, \dots, 9\}$ denote the digit in row $r$ and column $c$. 
Sudoku requires all digits within each row, column, and $3\times3$ box to be distinct. 
Writing $E$ for the set of unordered pairs of distinct cells sharing at least one of these units, the constraint is
\[
 \phi(\bz):=\bigwedge_{\{(r,c),(r',c')\}\in E}
 [z_{r,c}\neq z_{r',c'}].
\]
Each factor is a $9\times9$ Boolean tensor with zeros on the diagonal and ones elsewhere. 
This factorization introduces no auxiliary symbolic variables. 

\paragraph{ViSudo Completion}
For completion, clue-cell digits are the perceptual concepts $\bc$ and missing-cell digits are the output $\by$. 
The same constraint is evaluated on the combined board, giving $\phi(\bc,\by)$.

\paragraph{MNISTAdd}
Two $L$-digit numbers $A$ and $B$ are represented by $2L$ MNIST images, and the output is the $L+1$ digits of their sum $Y$. Indexing digits from least to most significant, the constraints describe the rules of column addition as:
\[
 A_0+B_0=Y_0+10C_0,\qquad
 A_i+B_i+C_{i-1}=Y_i+10C_i\quad(1\leq i<L),\qquad
 Y_L=C_{L-1}
\]
where $C_i$ are binary carries.
Each factor involves one column and its adjacent carries, giving a chain of local computations. 
The variable ordering of both SDD and \apricot follows this structure.

\paragraph{MNISTMul}
The inputs are again $2L$ MNIST images, and the output is the $2L$ digits of $A * B$. 
A coarse factorization would connect input digits and carries as 
\[
\sum_{i+j=k}A_iB_j +C_{k-1}=Y_k+10C_k,\qquad 0\leq k<2L,
\]
with $C_{-1}=C_{2L-1}=0$. 
However, in this setting, each factor still involves a large number of input digits which exceed our GPU memory at $L=3$ in our implementation.
We therefore derive a finer factorization that introduces partial products $P_{ij}=A_iB_j$ and computes each column sum $S_k=\sum_{i+j=k}P_{ij}$ through a chain of running partial sums. 
With these auxiliary variables and factors, the carry logic becomes
\[
 S_k+C_{k-1}=Y_k+10C_k,\qquad 0\leq k<2L.
\]
with $S_{2L-1}=0$.
Notice that this is not possible for the fuzzy baseline, where an independent reduction of the finer factorization has ineffective learning gradients. 
We therefore retain the coarser carry formulation for fuzzy (explaining why fuzzy runs out of memory at $L=3$).

\paragraph{CLEVR-Diff}
Each of the $m$ objects has a key $K_i$, predicted by the perception and consisting of the shape and color of the object. We use $D=3\times 8=24$ possible keys. To check whether the $m$ objects are different, we evaluate a conjunction of factors that indicate whether all keys differ:
\[
\phi(\mathbf{K}) := 
 \bigwedge_{1\leq i<j\leq m} K_i\neq K_j.
\]
This constraint consists of $\binom{m}{2}$ pairwise inequalities and requires no auxiliary symbolic variables.

\paragraph{CLEVR-CountDiff}
Here, we use keys consisting of the shape, material, and size of the object, with $D=3\times2\times2=12$ possibilities. 
The output is the number of distinct keys, $Y=|\{K_1,\ldots,K_m\}|$ in the scene. 
The factorization uses auxiliary binary variables $S_{v,i}$ that record whether key $v$ occurs among the first $i$ objects. Then, running counts $T_v$ accumulate the number of keys present:
\[
\begin{aligned}
 S_{v,0}&=0,&
 S_{v,i}&\leftrightarrow S_{v,i-1}\lor[K_i=v],\\
 T_{-1}&=0,&
 T_v&=T_{v-1}+S_{v,m},\qquad Y=T_{D-1},
\end{aligned}
\]
for $0\leq v<D$ and $1\leq i\leq m$. 

\subsubsection{Datasets}
\paragraph{Visual Sudoku}
Both tasks use $9\times9$ boards with observed digits represented by MNIST images. ViSudo-Classification uses ViSudo-PC~\cite{augustine_2022_visudopc}: the label indicates whether a board satisfies the Sudoku rules. Training uses only valid boards, while evaluation includes valid and invalid boards. ViSudo-Completion uses the SatNet dataset~\cite{wang_2019_satnet}, with the data-leak correction of \citet{chang2023assessingsatnet}. Observed clue cells provide the image inputs, and the digits in the remaining cells are the supervised targets. We use the benchmark training and test splits.

\paragraph{Arithmetic}
We construct examples separately from the MNIST training and test splits:
disjoint groups of $2L$ images are used for the input numbers, and their digit labels determine the arithmetic output.
This gives $\lfloor60{,}000/(2L)\rfloor$ training and $\lfloor10{,}000/(2L)\rfloor$ test examples (therefore, increasing the input length $L$ also reduces the number of available training examples). 
The generated datasets are shared across methods and run seeds. 

\paragraph{Scene-understanding}
We extract object crops from annotated CLEVR-Hans3 scenes~\cite{stammer_right_2021}, resize them to $32\times32$, and assemble sets of $m$ objects as newly constructed scenes.
For CLEVR-Diff, we sample distinct keys and, with probability $1/2$, replace one by a duplicate of another, balancing the two output classes. 
For CLEVR-CountDiff, keys are drawn independently and uniformly. 
Each key is instantiated with a randomly selected crop of that type.
Training and test examples use crops exclusively from their respective source splits. 
We generate 10,000 training and 2,000 test examples for each task size, using fixed datasets across methods and runs. 
Recombining crops permits larger object sets than the original scenes and keeps the per-object sampling distribution fixed across $m$.

\subsection{WMC evaluation}
\label{app:wmc_evaluation}

For each arithmetic and CLEVR circuit, we evaluate 64 random weightings, shared across methods using a fixed seed. In each draw, every non-auxiliary variable receives an independently sampled $\operatorname{Dirichlet}(\mathbf{1})$ weight vector, uniform over its probability simplex. These include input digits or object keys and, where represented explicitly, output digits or counts. Auxiliary variables receive unit weights and are summed out. No evidence is fixed: the query sums over all satisfying assignments, including all output values. CLEVR-Diff evaluates the all-different constraint without an explicit output variable. These random queries assess circuit fidelity rather than performance under learned neural predictions or observed-label evidence.

Let $\wmc_b$ and $\wmchat_b$ denote the reference and estimated WMC for draw $b$.
For weight draw $b$, we measure the absolute log-ratio error
\[
    e_b
    =
    \left|
    \log_{10}\frac{\wmchat_b}{\wmc_b}
    \right|.
\]
We report the median of the 64 resulting errors,
\[
    E_{\mathrm{WMC}}
    =
    \operatorname{median}\{e_1,\ldots,e_{64}\}.
\]
Thus, $E_{\mathrm{WMC}}=1$ means that, for a typical sampled weighting, the estimated WMC differs from the reference by a factor of $10$.

\paragraph{Exact references}
In tasks where exact compilation breaks in the harder settings (MNISTMul, CLEVR-Diff, CLEVR-CountDiff), the exact references $\wmc_b$ are derived using task-specific solutions developed separately for each benchmark.
These specialized solutions used dynamic programming, inclusion-exclusion, or brute-force enumeration; they are not general-purpose WMC algorithms and are not included as competing methods. 
The procedures were developed with the assistance of AI and then checked against exact SDD evaluation on the easier instances for which SDD compilation was feasible. 

\begin{itemize}
\item \textbf{MNISTMul.}
For every pair of $L$-digit input numbers $(A,B)$, we compute the unique product $Y=AB$. The contribution of this assignment is the product of the weights assigned to the digits of $A$, $B$, and $Y$. Summing these contributions over all $10^{2L}$ input pairs gives the exact WMC. This direct enumeration is practical for the evaluated instances, where $L\leq3$.

\item \textbf{CLEVR-Diff.}
The required WMC is the total weight of assignments in which all $m$ objects receive different keys. We process the $D$ possible key values one at a time. A dynamic-programming state records the subset $S$ of objects that have already been assigned distinct processed keys. For the next key, we either leave it unused or assign it to one object not in $S$, multiplying by that object's weight for the key. After all keys have been processed, the state containing all objects is the exact all-different WMC. The algorithm has $\cO(2^m)$ states and costs $\cO(Dm2^m)$.

\item \textbf{CLEVR-CountDiff.}
Here we require the weighted distribution of the number of distinct keys. For every subset $B$ of the $D$ possible key values, we first compute
\[
    q(B)
    =
    \prod_{i=1}^{m}
    \sum_{v\in B}p_i(v),
\]
the total weight of assignments in which every object chooses a key from $B$. These quantities include assignments using all of $B$ as well as assignments using only a smaller subset. Inclusion--exclusion removes this repeated counting and recovers the total weight of assignments using exactly $t$ distinct keys, for every $t$. The output weights are then applied to this exact distribution. 
The computation is exponential in the key-domain size $D$ and linear in the number of objects $m$, making it practical here because $D=12$.

\end{itemize}

\subsection{Training and evaluation details}
\label{app:traineval_details}

\paragraph{\apricot configurations}
The grid of \apricot configurations evaluated in the Sudoku compilation ablation is detailed in \cref{app:sudoku_ablation}.
For the Visual Sudoku learning experiments, the selected compilation stage uses dense pre-multiplication groups of at most $10^6$ entries, leaving 113 initial factors, an initial TT-SVD rank $30$, and a maximum rank of $500$ during multiplications. 
For arithmetic and CLEVR, the selected \apricot C configurations skip pre-multiplications and use \TTround for product recompression, with maximum ranks below of $500$ for all experiments, except MNISTMul where we allow a maximum rank of $1200$. 
The multiplication schedules are different in each experiment and detailed in \cref{app:partitioning_mul}.

\paragraph{Compute budgets}
The reported configuration uses one NVIDIA A10G GPU with approximately 22 GiB memory. 
TT compilation runs on the GPU. 
SDD compilation runs on the CPU process with a 64 GiB memory cap and a two-hour time limit for the compilation study. 
Training has a six-hour wall-clock limit. 
Runs exceeding these limits are reported as out of memory or timed out, without increasing the budget. 
Results are averaged over ten runs.

\paragraph{Learning objectives}
We maximize the WMC in \cref{eq:wmc} and conditional WMC in \cref{eq:conditional_wmc} by setting the weights of the concepts $\bc$ to predicted neural probabilities and the weights of the output variables $\by$ to the given one-hot labels, and by marginalizing the circuit.
Auxiliary variables receive unit weights.
In practice, we maximize the logarithm of the resulting WMC.

\paragraph{Inference}
In the conditional WMC setting (\cref{app:conditional_wmc}), two strategies can be used to predict the output $\by$ from the latent concepts $\bc$ at test time.
The first, \emph{greedy decoding}, selects the most likely concept prediction from the neural predictor $\operatorname{argmax}_\bc p(\bc|x)$. 
The concepts are then fed to a symbolic solver to evaluate the constraint or derive output symbols $\by$, e.g., by running a Sudoku solver on the partially filled concepts or computing ordinary exact addition/multiplication for arithmetic. 
Under this strategy, the WMC is not used at inference time; only during learning. 
We use greedy decoding for the Visual Sudoku experiments.
The second, \emph{reasoning-based decoding}, uses the WMC at inference to predict $\operatorname{argmax}_\by \sum_\bc p(\bc|x) \phi(\bc,\by)$ -- or doing this independently for each $y_k$.
This approach can pick up on subtle inconsistencies in the predictions and correct them based on the constraints. 
We use it for the arithmetic and CLEVR experiments. 

\paragraph{Architectures and hyperparameters}
We use conventional CNNs to process each input image independently (an MNIST image or an object crop) and produce a softmax distribution over its symbol.
Architecture details are provided in the repository.
We use Adam~\cite{kingma2014adam} with the following settings, together with usual techniques such as gradient clipping.
There is no validation-based selection or early stopping: metrics use the final epoch.

\begin{center}
\begin{tabular}{l r r r}
\toprule
Task & Learning rate & Batch size & Epochs\\
\midrule
MNISTAdd & $10^{-3}$ & 32 (8 for $L=100$) & 30\\
MNISTMul & $10^{-3}$ & 32 & 10\\
CLEVR-Diff / CountDiff & $10^{-3}$ & 32 & 40\\
ViSudo-Completion & $10^{-5}$ & 8 & 20\\
ViSudo-Classification & $10^{-3}$ & 20 & 2000\\
\bottomrule
\end{tabular}
\end{center}

\section{Approximation Algorithms for TT Multiplication}
\label{app:tt_mult_algos}

In this appendix, we detail the four recompression strategies introduced in
\cref{sec:tt_approx_mult}. All algorithms take as input two compatible TTs
$\bcA$ and $\bcB$ (same scope, same core ordering, same mode dimensions $n_k$)
and a target maximum rank $R$ or Frobenius tolerance $\epsilon$, and return a TT
approximation $\approxhat{\bcX}$ of the Hadamard product $\bcX = \bcA * \bcB$.
We denote by $\bcT_{\bcA,k} \in \R^{r^A_{k-1} \times n_k \times r^A_{k}}$ the $k$-th
core of $\bcA$, and similarly for $\bcB$ and $\approxhat{\bcX}$.

\paragraph{Notation} We follow einsum and einops conventions for tensor
operations~\cite{rogozhnikov2022einops}. Given tensors $\bcT_1, \bcT_2$:
\begin{itemize}
    \item $\textsc{einsum}(\texttt{"abc, cde -> abde"},\, \bcT_1, \bcT_2)$ denotes
    Einstein summation, contracting indices that appear in both inputs but not
    in the output.
    \item $\textsc{rearrange}(\texttt{"(a b) c -> a b c"},\, \bcT)$ denotes a
    reshape/permutation, where parentheses on the left split a single axis into
    multiple axes (with explicit sizes given as side arguments) and parentheses
    on the right merge multiple axes into a single axis.
\end{itemize}

\subsection{\TTround}
\label{app:ttround}

\TTround \cite{oseledets2011tensor} computes $\approxhat{\bcX} \approx \bcA * \bcB$
by interleaving the Kronecker product of cores with a sweep of truncated SVDs.
The naive variant first materializes the inflated cores
$\bcT_{\bcX,k} = \bcT_{\bcA,k} \otimes \bcT_{\bcB,k}$ of rank $r^A_k r^B_k$
and then applies a standard rounding sweep, paying $\cO(dnr^4)$ memory for the
inflated cores. The \emph{single-pass} variant presented here exploits the
Matrix-Kronecker structure of the inflated cores to avoid this materialization,
reducing the memory cost to $\cO(r^3)$.

The algorithm proceeds in two phases:
\begin{enumerate}
    \item \textbf{Right-orthogonalization.} Both $\bcA$ and $\bcB$ are first
    right-orthogonalized so that the right unfolding of every core is
    row-orthonormal. This ensures that the subsequent local SVD truncations are
    quasi-optimal with respect to the global Frobenius norm.
    \item \textbf{Left-to-right SVD sweep.} A left environment
    $\bcL \in \R^{r_{k-1} \times r^A_{k-1} \times r^B_{k-1}}$ accumulates the
    already-compressed prefix. At each site $k$, the algorithm (i) contracts
    $\bcL$ with the (implicit) Kronecker slice
    $\bcT_{\bcA,k} \otimes \bcT_{\bcB,k}$ via a single einsum, producing a
    working tensor $\bcW$; (ii) flattens $\bcW$ into a matrix and performs a
    truncated SVD with target rank $R$ (or tolerance $\epsilon$); (iii) stores
    the left singular vectors as the new core $\bcT_{\approxhat{\bcX},k}$ and
    absorbs $\mathrm{diag}(\bs)\bV^\top$ into the next left environment.
\end{enumerate}
The last core requires no SVD truncation since its rightmost rank is already
$1$ by definition.

\begin{algorithm}[t]
\caption{Hadamard \TTround}
\label{alg:hadamard_round}
\begin{algorithmic}[1]
\Require TTs $\bcA, \bcB$ with cores $\{\bcT_{\bcA,k}\}_{k=1}^d$, $\{\bcT_{\bcB,k}\}_{k=1}^d$; max rank $R$; tolerance $\epsilon$.
\Ensure TT approximation $\approxhat{\bcX} \approx \bcA * \bcB$.
\State $\bcA \gets \textsc{RightOrthogonalize}(\bcA)$
\State $\bcB \gets \textsc{RightOrthogonalize}(\bcB)$
\State $\bcL \gets \mathbf{1} \in \R^{1 \times 1 \times 1}$ \Comment{initial left environment}
\For{$k = 1, \ldots, d$}
    \State $\bcW \gets \textsc{einsum}\big(\texttt{"r ra rb,\ ra n ra',\ rb n rb' -> r n ra' rb'"},\ \bcL,\ \bcT_{\bcA,k},\ \bcT_{\bcB,k}\big)$
    \State $\bcW \gets \textsc{rearrange}\big(\texttt{"r n ra' rb' -> (r n) (ra' rb')"},\ \bcW\big)$
    \If{$k < d$}
        \State $\bU, \bs, \bV^\top \gets \textsc{TruncatedSVD}(\bcW;\ \epsilon,\ R)$
        \State $\bcT_{\approxhat{\bcX},k} \gets \textsc{rearrange}\big(\texttt{"(r n) r' -> r n r'"},\ \bU\big)$
        \State $\bcL \gets \textsc{rearrange}\big(\texttt{"r' (ra' rb') -> r' ra' rb'"},\ \mathrm{diag}(\bs)\,\bV^\top\big)$
    \Else
        \State $\bcT_{\approxhat{\bcX},d} \gets \textsc{rearrange}\big(\texttt{"(r n) 1 -> r n 1"},\ \bcW\big)$
    \EndIf
\EndFor
\State \Return TT $\approxhat{\bcX}$ with cores $\{\bcT_{\approxhat{\bcX},k}\}_{k=1}^d$.
\end{algorithmic}
\end{algorithm}

\paragraph{Randomized per-site SVD} When the target rank $R$ is small
relative to the working tensor dimensions, the truncated SVD on $\bcW$
(\cref{alg:hadamard_round}, line 8) can be replaced by a randomized low-rank SVD \cite{halko2010randomsvd}: $\bcW$ is multiplied by a Gaussian sketch of width $R + r_{\text{oversample}}$, and the dominant left singular vectors are recovered from the resulting tall-and-skinny matrix. 
We emphasize that this is a purely \emph{local}, matrix-level acceleration --- a fresh sketch is drawn independently at each site, and the sweep structure, orthogonalization, and per-site target ranks are unchanged. 
This is conceptually distinct from the TT-level sketching used by \HaTT (\cref{app:hatt}), which draws a single
random TT once and reuses it across all sites by precomputing right environments, so that no unfolding of $\bcX$ is ever formed.

\paragraph{Complexity} At site $k$, $\bcW$ has shape $r_{k-1} n_k \times r^A_k r^B_k$
and its truncated SVD costs $\cO(n r^4)$ for uniform ranks $r$. Summed over $d$
sites, this gives $\cO(d n r^4)$ time. Memory is dominated by storing $\bcW$
and the left environment, both of which are $\cO(r^3)$ since the Kronecker
factors are never explicitly formed. When randomized per-site SVDs are used in
place of deterministic ones, the per-site cost drops to $\cO(n r^3)$ since the
SVD operates on a sketched matrix of column dimension $\cO(r)$ rather than on
the full $r^A_k r^B_k = r^2$ unfolding, giving an overall time complexity of
$\cO(d n r^3)$ --- matching that of \HaTT (\cref{app:hatt}).

\subsection{\HaTT}
\label{app:hatt}

\HaTT \cite{sun2025hatt} replaces the deterministic SVD-based truncation of
\TTround with a randomized sketching approach inspired by randomized linear
algebra \cite{halko2010randomsvd,daas2021randomizedTT}. Like \TTround, it
exploits the Matrix-Kronecker structure of the inflated cores
$\bcT_{\bcA,k} \otimes \bcT_{\bcB,k}$ to avoid materializing them. 
Unlike \TTround --- whose per-site SVDs (deterministic or randomized) still operate on full unfoldings of the inflated working tensor $\bcW$ --- \HaTT projects $\bcX$ against a TT sketch $\bcS$ of width
$s = R + r_{\text{oversample}}$ first and applies SVDs only on small sketched matrices of column dimension $s$.

The algorithm proceeds in two phases.

\paragraph{Phase 1: Sketched right environments} A random TT sketch $\bcS$
with cores $\bcT_{\bcS,k} \in \R^{s_{k-1} \times n_k \times s_k}$ of width
$s$ is drawn (e.g., with Gaussian or Rademacher entries) and
left-orthogonalized for numerical stability. A right-to-left sweep
(\cref{alg:sketched_right_envs}) then produces the right environments
$\bcR_k \in \R^{r^A_k \times r^B_k \times s_k}$ for $k = 1, \ldots, d-1$.
Each $\bcR_k$ represents the contraction of the inflated cores at positions
$k+1, \ldots, d$ projected against the right tail of $\bcS$.

As an extension to the original \HaTT formulation, we additionally allow the
user to provide a TT $\bcG$ as an initial guess of the solution, which is then
used in place of a random sketch. This can leverage prior knowledge about
$\bcX$ -- for instance, the result of a previous Hadamard product in a
bottom-up compilation -- to bias the sketch towards the relevant subspace.

\paragraph{Phase 2: Left-to-right projection sweep} A left-to-right sweep
analogous to that of \TTround constructs the cores of $\approxhat{\bcX}$. At
each site $k$, the algorithm contracts the left environment $\bcL$, the
inflated core $\bcT_{\bcA,k} \otimes \bcT_{\bcB,k}$, and the sketched right
environment $\bcR_k$ into a small working tensor $\bM$ of shape
$r_{k-1} \times n_k \times s_k$, where $r_{k-1}$ is the running compressed
rank. A truncated SVD on the unfolding of $\bM$ then yields the left singular
vectors used as the new core $\bcT_{\approxhat{\bcX}, k}$. The left
environment is updated by contracting the new core back against the inflated
cores, and the loop proceeds. As in \TTround, the last core requires no SVD
since $s_d = 1$.

\begin{algorithm}[t]
\caption{Sketched right environments for $\bcX = \bcA * \bcB$}
\label{alg:sketched_right_envs}
\begin{algorithmic}[1]
\Require TT cores $\{\bcT_{\bcA,k}\}_{k=1}^d, \{\bcT_{\bcB,k}\}_{k=1}^d$; sketch TT cores $\{\bcT_{\bcS,k}\}_{k=1}^d$.
\Ensure Right environments $\{\bcR_k\}_{k=1}^{d-1}$.
\State $\bcR_d \gets \mathbf{1} \in \R^{1 \times 1 \times 1}$
\For{$k = d, d-1, \ldots, 2$}
    \State $\bcR_{k-1} \gets \begin{aligned}[t] \textsc{einsum}\big(&\texttt{"ra n ra',\ rb n rb',\ s n s',\ ra' rb' s' -> ra rb s"},\\ &\bcT_{\bcA,k},\ \bcT_{\bcB,k},\ \bcT_{\bcS,k},\ \bcR_k\big)\end{aligned}$
\EndFor
\State \Return $\{\bcR_k\}_{k=1}^{d-1}$
\end{algorithmic}
\end{algorithm}

\begin{algorithm}[t]
\caption{Hadamard \HaTT}
\label{alg:hatt}
\begin{algorithmic}[1]
\Require TTs $\bcA, \bcB$; max rank $R$; tolerance $\epsilon$; sketch width $s = R + r_{\text{oversample}}$; (optional) initial guess TT $\bcG$.
\Ensure TT approximation $\approxhat{\bcX} \approx \bcA * \bcB$.
\State $\bcA \gets \textsc{RightOrthogonalize}(\bcA)$
\State $\bcB \gets \textsc{RightOrthogonalize}(\bcB)$
\State $\bcS \gets \bcG$ if provided, else $\textsc{RandomTT}(\text{dims},\ s)$
\State $\bcS \gets \textsc{LeftOrthogonalize}(\bcS)$
\State $\{\bcR_k\}_{k=1}^{d-1} \gets \textsc{SketchedRightEnvs}(\bcA, \bcB, \bcS)$ \Comment{\cref{alg:sketched_right_envs}}
\State $\bcR_d \gets \mathbf{1} \in \R^{1 \times 1 \times 1}$
\State $\bcL \gets \mathbf{1} \in \R^{1 \times 1 \times 1}$ \Comment{initial left environment}
\For{$k = 1, \ldots, d$}
    \State $\bM \gets \begin{aligned}[t]\textsc{einsum}\big(&\texttt{"r ra rb,\ ra n ra',\ rb n rb',\ ra' rb' s' -> r n s'"},\\ & \bcL,\ \bcT_{\bcA,k},\ \bcT_{\bcB,k},\ \bcR_k\big)\end{aligned}$
    \If{$k < d$}
        \State $\bM \gets \textsc{rearrange}\big(\texttt{"r n s' -> (r n) s'"},\ \bM\big)$
        \State $\bU,\, \cdot,\, \cdot \gets \textsc{TruncatedSVD}(\bM;\ \epsilon,\ R)$
        \State $\bcT_{\approxhat{\bcX},k} \gets \textsc{rearrange}\big(\texttt{"(r n) r' -> r n r'"},\ \bU\big)$
        \State $\bcL \gets \begin{aligned}[t]\textsc{einsum}\big(&\texttt{"r n r',\ r ra rb,\ ra n ra',\ rb n rb' -> r' ra' rb'"},\\ & \bcT_{\approxhat{\bcX},k},\ \bcL,\ \bcT_{\bcA,k},\ \bcT_{\bcB,k}\big)\end{aligned}$
    \Else
        \State $\bcT_{\approxhat{\bcX},d} \gets \bM$ \Comment{$s_d = 1$, so $\bM$ already has shape $r_{d-1} \times n_d \times 1$}
    \EndIf
\EndFor
\State \Return TT $\approxhat{\bcX}$ with cores $\{\bcT_{\approxhat{\bcX},k}\}_{k=1}^d$.
\end{algorithmic}
\end{algorithm}

\paragraph{Complexity} The dominant cost at each site is the truncated SVD
on $\bM$ of shape $r_{k-1} n_k \times s_k$, costing $\cO(n r^2 s)$ under
uniform ranks. Summed over the $d$ sites, this gives $\cO(d n r^2 s)$ time.
Memory is dominated by storing the right environments, at $\cO(r^2 s)$ each.
The right-environment sweep (\cref{alg:sketched_right_envs}) has the same
asymptotic cost of $\cO(d n r^2 s)$. For $s \approx r$, the overall time
becomes $\cO(d n r^3)$.

\subsection{\TCI}
\label{app:tci}

Tensor Cross Interpolation (\TCI)
\cite{oseledets_tt-cross_2010,NunezFernandes2025TCI} approximates the Hadamard
product $\bcX = \bcA * \bcB$ by a TT $\approxhat{\bcX}$ that interpolates
$\bcX$ at a small set of carefully chosen entries. Unlike \TTround and \HaTT,
\TCI never materializes a Kronecker core nor performs an SVD on a full
unfolding of $\bcX$: at each site, a small \emph{supercore} is formed by
contracting a $w$-site window of the input cores with their interpolated
environments, and an interpolative decomposition picks the rows or columns of
that supercore that best interpolate it. We present the algorithm with two-site
sweeps ($w = 2$); the generalization to $K > 2$ factors and larger windows is
straightforward.

\paragraph{Pivots and shared environments} Let $p \le R$ denote the current
number of pivots at a given bond. \TCI maintains per-factor environments that
share the pivot axis across factors:
\begin{itemize}
    \item Left environments
    $\bcL^{\bcA}_i \in \R^{p \times r^A_{i-1}}$ and
    $\bcL^{\bcB}_i \in \R^{p \times r^B_{i-1}}$ holding $p$ contracted slices
    of the left tail of $\bcA$ and $\bcB$ respectively.
    \item Right environments
    $\bcR^{\bcA}_i \in \R^{r^A_i \times p}$ and
    $\bcR^{\bcB}_i \in \R^{r^B_i \times p}$ holding $p$ contracted slices of
    the right tails.
\end{itemize}
The shared pivot axis encodes the Hadamard product structure without ever
forming Kronecker cores: at each bond, both factors are interpolated at the
\emph{same} set of physical-pivot indices.

\paragraph{Interpolative primitives} We use two primitives for selecting
pivots and interpolating from them.
\begin{itemize}
    \item \textsc{MaxVol} \cite{goreinov2010maxvol}: given a matrix $\bM$,
    $\textsc{MaxVol}(\bM; R)$ returns up to $R$ column indices of $\bM$ that
    form a near-maximum-volume submatrix.
    \item \textsc{prrLDU} \cite{savostyanov2014quasioptimality}: a partially
    rank-revealing LU decomposition. Its row-pivoted variant
    $\textsc{prrLDU}_\text{row}(\bM; \epsilon, R)$ returns an interpolation
    matrix $\bcZ$ and a pivot set such that
    $\bM \approx \bcZ \cdot \bM[\text{pivot}, :]$. The column-pivoted variant
    $\textsc{prrLDU}_\text{col}$ analogously yields
    $\bM \approx \bM[:, \text{pivot}] \cdot \bcZ$.
\end{itemize}

\paragraph{Phase 1: Maxvol warm-up} A right-to-left sweep
(\cref{alg:tci_maxvol}) initializes the right environments. At each site,
per-factor supercores $\bcPhi_{\bcA}, \bcPhi_{\bcB}$ are formed by
contracting the current right environment with the corresponding core; their
joint Hadamard matrix $\bcPhi$ has rows that range over the outer product
of the per-factor left bonds and columns that range over the shared
physical-pivot indices. \textsc{MaxVol} selects up to $R$ such columns, and
the picked columns of each $\bcPhi_{\bcA}, \bcPhi_{\bcB}$ become the
next right environments.

\paragraph{Phase 2: Alternating sweeps} Sweeps of \cref{alg:tci_l2r} (L2R)
and \cref{alg:tci_r2l} (R2L) refine the pivots and emit the cores of
$\approxhat{\bcX}$. At each interior site of an L2R sweep, per-factor
supercores are formed by contracting the left environment, two consecutive
cores, and the right environment. They are combined element-wise into the
joint supercore $\bcPhi$, flattened into a matrix, and decomposed by
$\textsc{prrLDU}_\text{row}$ into an interpolation matrix $\bcZ$ and a row
pivot. The interpolation matrix becomes the new core
$\bcT_{\approxhat{\bcX}, i}$, and the pivot updates the left environments of
$\bcA$ and $\bcB$. The R2L sweep is symmetric, processing sites from right
to left and using $\textsc{prrLDU}_\text{col}$ to update right environments.

\paragraph{Boundary cores} The L2R sweep additionally extracts the rightmost
core $\bcT_{\approxhat{\bcX}, d}$ directly from the boundary supercore at
$i = d-1$ (whose right rank is already $r_d = 1$). The R2L sweep mirrors this
for the leftmost core.

\begin{algorithm}[t]
\caption{Maxvol right environments for $\bcX = \bcA * \bcB$}
\label{alg:tci_maxvol}
\begin{algorithmic}[1]
\Require TTs $\bcA, \bcB$; max rank $R$.
\Ensure Right environments $\{\bcR^{\bcA}_i, \bcR^{\bcB}_i\}_{i=1}^{d-1}$.
\State $\bcR^{\bcA}_d \gets \mathbf{1} \in \R^{1 \times 1}$;\quad $\bcR^{\bcB}_d \gets \mathbf{1} \in \R^{1 \times 1}$
\For{$i = d, d-1, \ldots, 2$}
    \For{$\bcY \in \{\bcA, \bcB\}$}
        \State $\bcPhi_{\bcY} \gets \textsc{einsum}\big(\texttt{"r n r',\ r' p -> r (n p)"},\ \bcT_{\bcY,i},\ \bcR^{\bcY}_i\big)$
    \EndFor
    \State $\bcPhi \gets \textsc{einsum}\big(\texttt{"ra q,\ rb q -> (ra rb) q"},\ \bcPhi_{\bcA},\ \bcPhi_{\bcB}\big)$
    \State $\text{pivot} \gets \textsc{MaxVol}(\bcPhi;\ R)$
    \For{$\bcY \in \{\bcA, \bcB\}$}
        \State $\bcR^{\bcY}_{i-1} \gets \bcPhi_{\bcY}[:, \text{pivot}]$
    \EndFor
\EndFor
\State \Return $\{\bcR^{\bcA}_i, \bcR^{\bcB}_i\}_{i=1}^{d-1}$
\end{algorithmic}
\end{algorithm}

\begin{algorithm}[t]
\caption{Left-to-right \TCI sweep}
\label{alg:tci_l2r}
\begin{algorithmic}[1]
\Require TTs $\bcA, \bcB$; left envs $\{\bcL^{\bcA}_i, \bcL^{\bcB}_i\}_{i=1}^{d-1}$; right envs $\{\bcR^{\bcA}_i, \bcR^{\bcB}_i\}_{i=1}^{d}$; max rank $R$; tolerance $\epsilon$.
\Ensure Cores $\{\bcT_{\approxhat{\bcX}, i}\}_{i=1}^d$ and updated left envs.
\State $\bcL^{\bcA}_1 \gets \mathbf{1} \in \R^{1 \times 1}$;\quad $\bcL^{\bcB}_1 \gets \mathbf{1} \in \R^{1 \times 1}$
\For{$i = 1, \ldots, d-1$}
    \For{$\bcY \in \{\bcA, \bcB\}$}
        \State $\bcPhi_{\bcY} \gets \begin{aligned}[t]\textsc{einsum}\big( &\texttt{"p r,\ r n r',\ r' m r'',\ r'' q -> p n m q"},\\ & \bcL^{\bcY}_i,\ \bcT_{\bcY,i},\ \bcT_{\bcY,i+1},\ \bcR^{\bcY}_{i+1}\big)\end{aligned}$
    \EndFor
    \State $\bcPhi \gets \bcPhi_{\bcA} * \bcPhi_{\bcB}$ \Comment{element-wise Hadamard}
    \State $\bcPhi \gets \textsc{rearrange}\big(\texttt{"p n m q -> (p n) (m q)"},\ \bcPhi\big)$
    \State $(\bcZ,\ \text{pivot}) \gets \textsc{prrLDU}_\text{row}(\bcPhi;\ \epsilon,\ R)$
    \State $\bcT_{\approxhat{\bcX}, i} \gets \textsc{rearrange}\big(\texttt{"(p n) r' -> p n r'"},\ \bcZ\big)$
    \For{$\bcY \in \{\bcA, \bcB\}$}
        \State $\bcL^{\bcY}_{i+1} \gets \textsc{rearrange}\big(\textsc{einsum}\big(\texttt{"p r,\ r n r' -> (p n) r'"},\ \bcL^{\bcY}_i,\ \bcT_{\bcY,i}\big)\big)[\text{pivot}, :]$
    \EndFor
    \If{$i = d-1$}
        \State $\bcT_{\approxhat{\bcX}, d} \gets \textsc{rearrange}\big(\texttt{"r' n -> r' n 1"},\ \bcPhi[\text{pivot}, :]\big)$
    \EndIf
\EndFor
\State \Return cores $\{\bcT_{\approxhat{\bcX}, i}\}_{i=1}^d$, updated $\{\bcL^{\bcA}_i, \bcL^{\bcB}_i\}_{i=1}^d$
\end{algorithmic}
\end{algorithm}

\begin{algorithm}[t]
\caption{Right-to-left \TCI sweep}
\label{alg:tci_r2l}
\begin{algorithmic}[1]
\Require TTs $\bcA, \bcB$; left envs $\{\bcL^{\bcA}_i, \bcL^{\bcB}_i\}_{i=1}^{d}$; right envs $\{\bcR^{\bcA}_i, \bcR^{\bcB}_i\}_{i=1}^{d-1}$; max rank $R$; tolerance $\epsilon$.
\Ensure Cores $\{\bcT_{\approxhat{\bcX}, i}\}_{i=1}^d$ and updated right envs.
\State $\bcR^{\bcA}_d \gets \mathbf{1}$;\quad $\bcR^{\bcB}_d \gets \mathbf{1}$
\For{$i = d, d-1, \ldots, 2$}
    \For{$\bcY \in \{\bcA, \bcB\}$}
        \State $\bcPhi_{\bcY} \gets \begin{aligned}[t] \textsc{einsum}\big(&\texttt{"p r,\ r n r',\ r' m r'',\ r'' q -> p n m q"}, \\ 
            &\bcL^{\bcY}_{i-1},\ \bcT_{\bcY,i-1},\ \bcT_{\bcY,i},\ \bcR^{\bcY}_i\big)\end{aligned}$
    \EndFor
    \State $\bcPhi \gets \bcPhi_{\bcA} * \bcPhi_{\bcB}$
    \State $\bcPhi \gets \textsc{rearrange}\big(\texttt{"p n m q -> (p n) (m q)"},\ \bcPhi\big)$
    \State $(\bcZ,\ \text{pivot}) \gets \textsc{prrLDU}_\text{col}(\bcPhi;\ \epsilon,\ R)$
    \State $\bcT_{\approxhat{\bcX}, i} \gets \textsc{rearrange}\big(\texttt{"r' (m q) -> r' m q"},\ \bcZ\big)$
    \For{$\bcY \in \{\bcA, \bcB\}$}
        \State $\bcR^{\bcY}_{i-1} \gets \textsc{rearrange}\big(\textsc{einsum}\big(\texttt{"r m r',\ r' q -> r (m q)"},\ \bcT_{\bcY,i},\ \bcR^{\bcY}_i\big)\big)[:, \text{pivot}]$
    \EndFor
    \If{$i = 2$}
        \State $\bcT_{\approxhat{\bcX}, 1} \gets \textsc{rearrange}\big(\texttt{"n r' -> 1 n r'"},\ \bcPhi[:, \text{pivot}]\big)$
    \EndIf
\EndFor
\State \Return cores $\{\bcT_{\approxhat{\bcX}, i}\}_{i=1}^d$, updated $\{\bcR^{\bcA}_i, \bcR^{\bcB}_i\}_{i=1}^d$
\end{algorithmic}
\end{algorithm}

\begin{algorithm}[t]
\caption{\TCI for $\bcX = \bcA * \bcB$}
\label{alg:tci}
\begin{algorithmic}[1]
\Require TTs $\bcA, \bcB$; max rank $R$; tolerance $\epsilon$; number of sweeps $T$.
\Ensure TT approximation $\approxhat{\bcX} \approx \bcA * \bcB$.
\State $\bcA \gets \textsc{RightOrthogonalize}(\bcA)$;\quad $\bcB \gets \textsc{RightOrthogonalize}(\bcB)$
\State $\{\bcR^{\bcA}_i, \bcR^{\bcB}_i\} \gets \textsc{MaxvolRightEnvs}(\bcA, \bcB; R)$ \Comment{\cref{alg:tci_maxvol}}
\For{$t = 1, \ldots, T$}
    \State $(\text{cores},\ \{\bcL^{\bcA}_i, \bcL^{\bcB}_i\}) \gets \textsc{L2R-TCI}(\bcA, \bcB, \{\bcL\}, \{\bcR\}; R, \epsilon)$ \Comment{\cref{alg:tci_l2r}}
    \State $(\text{cores},\ \{\bcR^{\bcA}_i, \bcR^{\bcB}_i\}) \gets \textsc{R2L-TCI}(\bcA, \bcB, \{\bcL\}, \{\bcR\}; R, \epsilon)$ \Comment{\cref{alg:tci_r2l}}
\EndFor
\State \Return TT $\approxhat{\bcX}$ with cores $\text{cores}$
\end{algorithmic}
\end{algorithm}

\paragraph{Complexity} At each interior site, the dominant cost is forming the
joint supercore $\bcPhi$ and applying \textsc{prrLDU} to its
$(p \cdot n) \times (n \cdot p)$ unfolding. The per-factor four-tensor
contractions cost $\cO(n^2 p r^2)$, the elementwise Hadamard costs
$\cO(n^2 p^2)$, and \textsc{prrLDU} costs $\cO(n^2 p^3)$ in the worst case.
With $p \leq R \approx r$, the per-site cost is $\cO(n^2 r^3)$. Summing over
the $d$ sites and a constant number of sweeps gives an overall time complexity
of $\cO(d n^2 r^3)$ and memory $\cO(n^2 r^2)$, dominated by $\bcPhi$. The
maxvol warm-up has lower per-site cost, $\cO(n r^3)$, and is dominated by the
sweep phase.

\subsection{\RSI}
\label{app:rsi}

Randomized Sketching with Interpolation (\RSI) \cite{meng2026RSI} replaces
\TCI's maxvol-based right environment construction with random sketching.
Like \HaTT, \RSI sketches each input TT independently rather than evaluating
exact entries; like \TCI, it then performs interpolative-decomposition sweeps
to construct $\approxhat{\bcX}$. Compared to \HaTT, \RSI differs in two ways:
(i) it uses a Khatri-Rao product (KRP) sketch rather than a TT sketch, and
(ii) it uses interpolative decompositions (prrLDU) rather than SVDs to
truncate at each site. Because the sketching is per-factor, \RSI side-steps
the combinatorial maxvol search of \TCI (which is specific to the Hadamard
product) and immediately extends to any non-linear operation over multiple
TTs.

The original \RSI formulation uses a single L2R sweep after sketching. We
extend it to alternating L2R/R2L sweeps, which follow naturally from the \TCI
connection: once a first L2R sweep produces interpolative left environments,
subsequent sweeps refine the pivots in both directions exactly as in \TCI.

\paragraph{KRP sketches} The original \RSI paper uses a Khatri-Rao product
sketch: a set of $d$ matrices $\{\bcS_i \in \R^{n_i \times s}\}_{i=1}^d$ with
i.i.d. random entries (e.g., Gaussian or Rademacher), where
$s = R + r_\text{oversample}$ is the sketch width. Unlike the TT sketch used in
\HaTT --- whose sketch cores have an internal bond structure
$\bcT_{\bcS, k} \in \R^{s_{k-1} \times n_k \times s_k}$ --- the KRP sketch has
a single ``flat'' sketch axis $s$ shared across all $d$ sites. This makes the
contractions in the right-env sweep structurally simpler.

\paragraph{Phase 1: Sketched right environments} A right-to-left sweep
(\cref{alg:rsi_krp_envs}) computes the per-TT right environments
$\bcR^{\bcY}_i \in \R^{r^Y_i \times s}$ for $\bcY \in \{\bcA, \bcB\}$. At each
site, the previous right environment is updated by contracting the core with
the per-site sketch matrix on the physical dimension and broadcasting on the
shared sketch axis.

\paragraph{Phase 2: Alternating sweeps} The L2R and R2L sweeps of \TCI
(\cref{alg:tci_l2r,alg:tci_r2l}) are reused unchanged: the sketched right
environments play the same role in the supercore contractions as TCI's
maxvol-selected ones. After the first L2R sweep, the left environments become
interpolative (subset-based) just as in \TCI; from then on, the algorithm is
identical to \TCI. The original \RSI corresponds to stopping after this first
L2R sweep, before any R2L.

\begin{algorithm}[t]
\caption{KRP-sketched right environments for a TT $\bcA$}
\label{alg:rsi_krp_envs}
\begin{algorithmic}[1]
\Require TT cores $\{\bcT_{\bcA, i}\}_{i=1}^d$; KRP sketch matrices $\{\bcS_i \in \R^{n_i \times s}\}_{i=1}^d$.
\Ensure Right environments $\{\bcR^{\bcA}_i\}_{i=1}^{d-1}$.
\State $\bcR^{\bcA}_d \gets \mathbf{1} \in \R^{1 \times s}$ \Comment{all-ones vector along the sketch axis}
\For{$i = d, d-1, \ldots, 2$}
    \State $\bcR^{\bcA}_{i-1} \gets \textsc{einsum}\big(\texttt{"r n r',\ n s,\ r' s -> r s"},\ \bcT_{\bcA,i},\ \bcS_i,\ \bcR^{\bcA}_i\big)$
\EndFor
\State \Return $\{\bcR^{\bcA}_i\}_{i=1}^{d-1}$
\end{algorithmic}
\end{algorithm}

\begin{algorithm}[t]
\caption{\RSI for $\bcX = \bcA * \bcB$}
\label{alg:rsi}
\begin{algorithmic}[1]
\Require TTs $\bcA, \bcB$; max rank $R$; tolerance $\epsilon$; sketch width $s = R + r_\text{oversample}$; number of sweeps $T \geq 1$.
\Ensure TT approximation $\approxhat{\bcX} \approx \bcA * \bcB$.
\State $\bcA \gets \textsc{RightOrthogonalize}(\bcA)$;\quad $\bcB \gets \textsc{RightOrthogonalize}(\bcB)$
\State Draw KRP sketches $\{\bcS_i \in \R^{n_i \times s}\}_{i=1}^d$ with i.i.d. random entries
\State $\{\bcR^{\bcA}_i\} \gets \textsc{KRP-RightEnvs}(\bcA, \{\bcS_i\})$ \Comment{\cref{alg:rsi_krp_envs}}
\State $\{\bcR^{\bcB}_i\} \gets \textsc{KRP-RightEnvs}(\bcB, \{\bcS_i\})$
\For{$t = 1, \ldots, T$}
    \State $(\text{cores},\ \{\bcL^{\bcA}_i, \bcL^{\bcB}_i\}) \gets \textsc{L2R-TCI}(\bcA, \bcB, \{\bcL\}, \{\bcR\}; R, \epsilon)$ \Comment{\cref{alg:tci_l2r}}
    \State $(\text{cores},\ \{\bcR^{\bcA}_i, \bcR^{\bcB}_i\}) \gets \textsc{R2L-TCI}(\bcA, \bcB, \{\bcL\}, \{\bcR\}; R, \epsilon)$ \Comment{\cref{alg:tci_r2l}; original \RSI omits this and stops after the first L2R}
\EndFor
\State \Return TT $\approxhat{\bcX}$ with cores $\text{cores}$
\end{algorithmic}
\end{algorithm}

\paragraph{Complexity} Forming the KRP-sketched right environments costs
$\cO(d n r^2 s)$ per TT, dominated by the per-site einsum. The subsequent
\TCI sweeps inherit the per-sweep complexity of $\cO(d n^2 r^3)$ in time and
$\cO(n^2 r^2)$ in memory. With $s \approx r$ and a constant number of sweeps,
the overall cost matches \TCI: $\cO(d n^2 r^3)$ time, $\cO(n^2 r^2)$ memory.
The savings over \TCI come not from asymptotic complexity but from avoiding
the combinatorial maxvol search in the warm-up; the savings over \HaTT come
from per-site sketches replacing TT sketches in the right-env sweep.

\clearpage
\newpage
\section*{NeurIPS Paper Checklist}

\begin{enumerate}

\item {\bf Claims}
    \item[] Question: Do the main claims made in the abstract and introduction accurately reflect the paper's contributions and scope?
    \item[] Answer: \answerYes{}
    \item[] Justification: The contributions are listed explicitly at the end of the Introduction. The framework and tensor-train methods are developed in \cref{sec:background,sec:tensor_networks,sec:pipeline}, and the inference and learning claims are evaluated in \cref{sec:experiments}. 
    In particular, \cref{sec:learning_results} distinguishes tasks where independence reduction remains competitive from tasks with shared auxiliary variables, where preserving dependencies through compilation substantially improves learning.

\item {\bf Limitations}
    \item[] Question: Does the paper discuss the limitations of the work performed by the authors?
    \item[] Answer: \answerYes{}
    \item[] Justification: A dedicated ``Limitations'' paragraph follows the Conclusions and discusses the suitability of current recompression methods for sparse Boolean tensors as well as the need to study the effect of program factorization and auxiliary-variable structure across more neurosymbolic methods.

\item {\bf Theory assumptions and proofs}
    \item[] Question: For each theoretical result, does the paper provide the full set of assumptions and a complete (and correct) proof?
    \item[] Answer: \answerYes{}
    \item[] Justification: Theoretical results (e.g. \cref{thm:eps_mult} and the corresponding corollary) are stated with all assumptions and proved in \cref{app:error_bounds}; supporting derivations for the independence approximation are given in \cref{app:indep_categorical_binary}.

    \item {\bf Experimental result reproducibility}
    \item[] Question: Does the paper fully disclose all the information needed to reproduce the main experimental results of the paper to the extent that it affects the main claims and/or conclusions of the paper (regardless of whether the code and data are provided or not)?
    \item[] Answer: \answerYes{}
    \item[] Justification: We describe the framework and pipeline choices in \cref{sec:pipeline}; provide tensor-train operations, multiplication schedules, and recompression algorithms in \cref{app:tt_algebra,app:partitioning,app:tt_broadcasting,app:tt_mult_algos}; define the six learning tasks and their factorizations in \cref{app:programs}; and report compilation settings, compute budgets, learning objectives, decoding procedures, and hyperparameters in \cref{app:traineval_details}.

\item {\bf Open access to data and code}
    \item[] Question: Does the paper provide open access to the data and code, with sufficient instructions to faithfully reproduce the main experimental results, as described in supplemental material?
    \item[] Answer: \answerYes{}
    \item[] Justification: Code and instructions for reproducing the experiments are available through the public repository linked in the paper. Dependencies are pinned and reproducibly installable with the \texttt{uv} package manager.

\item {\bf Experimental setting/details}
    \item[] Question: Does the paper specify all the training and test details (e.g., data splits, hyperparameters, how they were chosen, type of optimizer) necessary to understand the results?
    \item[] Answer: \answerYes{}
    \item[] Justification: \Cref{sec:experiments} describes the benchmarks, baselines, metrics, and main experimental setup. \Cref{app:programs,app:wmc_evaluation,app:traineval_details} specify the task construction and splits, weighted-inference protocol, compilation configurations, optimizer, learning rate, batch size, number of epochs, and decoding procedure.

\item {\bf Experiment statistical significance}
    \item[] Question: Does the paper report error bars suitably and correctly defined or other appropriate information about the statistical significance of the experiments?
    \item[] Answer: \answerYes{}
    \item[] Justification: Results are averaged over ten runs with fixed datasets across methods and runs, capturing variations across training and compute randomness. Figures report the mean and standard-deviation error bars across runs.

\item {\bf Experiments compute resources}
    \item[] Question: For each experiment, does the paper provide sufficient information on the computer resources (type of compute workers, memory, time of execution) needed to reproduce the experiments?
    \item[] Answer: \answerYes{}
    \item[] Justification: \Cref{app:traineval_details} specifies the NVIDIA A10G GPU and its memory, the CPU memory cap for SDD compilation, and the compilation and training wall-clock limits. Compilation times are reported in the inference figures, and runs exceeding the stated time or memory budgets are marked as timed out or out of memory.
    
\item {\bf Code of ethics}
    \item[] Question: Does the research conducted in the paper conform, in every respect, with the NeurIPS Code of Ethics \url{https://neurips.cc/public/EthicsGuidelines}?
    \item[] Answer: \answerYes{}
    \item[] Justification: We have reviewed the NeurIPS Code of Ethics and the research conforms to it; the work involves no human subjects, no personal data, and does not pose ethical concerns.

\item {\bf Broader impacts}
    \item[] Question: Does the paper discuss both potential positive societal impacts and negative societal impacts of the work performed?
    \item[] Answer: \answerNA{}
    \item[] Justification: The paper is foundational research on tractable inference primitives for neurosymbolic systems, with no direct path to positive or negative societal applications beyond the general impact of more reliable probabilistic reasoning tools.
    
\item {\bf Safeguards}
    \item[] Question: Does the paper describe safeguards that have been put in place for responsible release of data or models that have a high risk for misuse (e.g., pre-trained language models, image generators, or scraped datasets)?
    \item[] Answer: \answerNA{}
    \item[] Justification: The paper releases no pre-trained generative models, image generators, or scraped datasets with risk of misuse.

\item {\bf Licenses for existing assets}
    \item[] Question: Are the creators or original owners of assets (e.g., code, data, models), used in the paper, properly credited and are the license and terms of use explicitly mentioned and properly respected?
    \item[] Answer: \answerYes{}
    \item[] Justification: We cite the original works for all datasets, baselines, and external libraries used (e.g., ViSudo datasets, SatNet, KaHyPar, NetworkX), and use them under their standard research licenses.

\item {\bf New assets}
    \item[] Question: Are new assets introduced in the paper well documented and is the documentation provided alongside the assets?
    \item[] Answer: \answerNA{}
    \item[] Justification: The paper does not release new datasets or pre-trained models. The code is research code provided with reproducibility instructions but does not consist in an asset.

\item {\bf Crowdsourcing and research with human subjects}
    \item[] Question: For crowdsourcing experiments and research with human subjects, does the paper include the full text of instructions given to participants and screenshots, if applicable, as well as details about compensation (if any)?
    \item[] Answer: \answerNA{}
    \item[] Justification: The paper involves no crowdsourcing or research with human subjects.

\item {\bf Institutional review board (IRB) approvals or equivalent for research with human subjects}
    \item[] Question: Does the paper describe potential risks incurred by study participants, whether such risks were disclosed to the subjects, and whether Institutional Review Board (IRB) approvals (or an equivalent approval/review based on the requirements of your country or institution) were obtained?
    \item[] Answer: \answerNA{}
    \item[] Justification: The paper involves no human subjects research.

\item {\bf Declaration of LLM usage}
    \item[] Question: Does the paper describe the usage of LLMs if it is an important, original, or non-standard component of the core methods in this research? Note that if the LLM is used only for writing, editing, or formatting purposes and does \emph{not} impact the core methodology, scientific rigor, or originality of the research, declaration is not required.
    \item[] Answer: \answerYes{}
    \item[] Justification: We include a voluntary AI use statement in the acknowledgements. LLMs are not part of the core methodology of this paper. Use of generative AI tools was limited to help in coding some experiments, not the model, and was reviewed by the authors.

\end{enumerate}

\end{document}